\documentclass[preprint,12pt]{elsarticle}

\usepackage{amsmath,epsfig}
\usepackage{amssymb,amsfonts,amsthm,bm}
\usepackage{pstricks}
\usepackage{subcaption}
\usepackage{makecell}
\usepackage{color}
\usepackage[numbers]{natbib}
\usepackage{etoolbox}
\usepackage{xstring}
\usepackage{url}

\newrgbcolor{green}{0 1 0}
\newrgbcolor{darkgreen}{0 .65 0}

\begin{document}

\begin{frontmatter}



\title {Color Texture Image Retrieval Based on Copula Multivariate Modeling in the Shearlet Domain}


\author{Sadegh~Etemad,
        Maryam~Amirmazlaghani}
\address{{\small Amirkabir University of Technology, Department of Computer Engineering}\\
{\tt etemad.sadegh@aut.ac.ir}
}
\address{ {\small Amirkabir University of Technology, Department of Computer Engineering}\\
{\small P.O. Box 15914, Tehran, Iran}\\
{\tt mazlaghani@aut.ac.ir}\\
{\small Tel: 98 21 64542704, Fax: 98 21 66406469}}

\def\x{{\mathbf x}}
\def\L{{\cal L}}

\begin{abstract}
	\bf In this paper, a color texture image retrieval framework is proposed based on Shearlet domain modeling using Copula multivariate model. In the proposed framework, Gaussian Copula is used to model the dependencies between different sub-bands of the Non-Subsample Shearlet Transform (NSST) and non-Gaussian models are used for marginal modeling of the coefficients. Six different schemes are proposed for modeling NSST coefficients based on the four types of neighboring defined; moreover, Kullback–Leibler Divergence(KLD) close form is calculated in different situations for the two Gaussian Copula and non-Gaussian functions in order to investigate the similarities in the proposed retrieval framework. The Jeffery divergence (JD) criterion, which is a symmetrical version of KLD, is used for investigating similarities in the proposed framework. We have implemented our experiments on four texture image retrieval benchmark datasets, the results of which show the superiority of the proposed framework over the existing state-of-the-art methods. In addition, the retrieval time of the proposed framework is also analyzed in the two steps of feature extraction and similarity matching, which also shows that the proposed framework enjoys an appropriate retrieval time.\\
\end{abstract}

\begin{keyword}
Color Texture Image Retrieval, Shearlet Transform, Multivariate Copula, Jeffery Divergence.
\end{keyword}

\end{frontmatter}

\section{INTRODUCTION AND RELATED WORKS}

\subsection{INTRODUCTION}
\indent Nowadays, with the advancement of technology in the imaging industry, we are faced with a great number of recorded images. With the increasing number of images available on the internet, intelligent image-based search has become an important issue. Content-based image retrieval (CBIR) is one of the applications of image processing that could be used to solve this issue. The purpose of CBIR is to find among a very large number of images, a number of them with contents identical to the query image. One of the basic issues in the field of CBIR, is the retrieval of the texture in the images. Each of the components of images, include textures, the correct and accurate retrieval of which could be used in accurately retrieving the full image.
\\
\indent Basically, retrieving textural images contains two main steps: 1) Feature extraction and modeling step, where the appropriate features are extracted from the images in the database and the query image, and then the suitable features are selected among them and the modeling is executed. 2) Similarity measurement (SM), where a set of images in the database most similar to the query image are searched for based on the features extracted in the previous step. Achieving a accurate image retrieval system with proper speed, depends on wise selection in the two aforementioned steps. Since the feature extraction step is considered as the first step in designing an image retrieval system, it needs to be further explored. Usually in the field of feature extraction, the methods are trying to obtain High Level Semantic Information (HLSI) with the help of Low-Level Features (LLFs) \cite{Celik}. Among the low-level features are features such as color, texture, edge, and the shapes in the image that can be extracted using a variety of methods. Extraction of these features could be done at two global (all of the image) or local (an area in the image) scales \cite{Dharani}. 
\\
\indent Some studies show that the human visual system divides an object into smaller pieces and performs classification and image retrieval based on local information \cite{Landy}. That is why the use of local descriptors (local features) has received a lot of attention among the researchers in the field of machine vision. Based on \cite{Cao}, the local descriptors are divided into three main categories: 1) distribution-based, 2) spatial-frequency, 3) descriptors other than the two above.\\
\indent The first category of descriptors, display the appearance features of the images using mean histograms of areas in the image. Among the methods in this category are Local Binary Pattern (LBP), Scale Invariant Feature Transform (SIFT), Speeded Up Robust Features (SURF), and Histogram of Oriented Gradients (HOG). In the second category, frequency-spatial transforms are used as a means to create descriptors. Wavelet Transform (WT), Gabor Wavelet Transform (GWT), Dual-Tree Complex Wavelet Transform (DT-CWT), Contourlet Transform (CT), and Curvelet Transform are among the most important transforms in this category. In the third category, a set of image derivatives and moments which are known as descriptors, are calculated for approximating the image. Among the most important methods in the third category, are Generalized Moment Invariant (GMI), and illumination invariant multiscale auto-convolution moments. \\
\indent In addition to the local features that have caught the attention of researchers, Sparse Representation (SR) is also of great importance. Recently, in order to design an image retrieval system, as well as considering sparsity, researchers have mostly turned to the second category of descriptors \cite{Li2017,Yang,Karine,Ghodhbani,Li2019,Yildizer,ElAlami}. In the following, the most prominent studies in the field of image retrieval are introduced, taking into account the local descriptors from the second category.  
\\

\subsection{RELATED WORKS}
\indent In this paper, we examine the prominent works in the literature in the field of image retrieval using frequency transform descriptors from two perspectives, namely the type of extracted feature and the similarity criterion. \\
Reference \cite{Kwitt2009} could be mentioned as one of first outstanding studies that used descriptors based on spatial-frequency transforms in image retrieval. Kwitt et al. \cite{Kwitt2009}, modeled DT-CWT coefficients in order to retrieve color texture images. To this end, they have performed joint modeling using the student t Copula and the Weibull distribution functions. Finally, they examine the similarity between the query image and distribution of the candid images using KLD criterion. It is stated there that joint modeling of the coefficients improves the image retrieval rates. Following their previous work, in 2010 Kwitt et al. \cite{Kwitt2009J} proposed a lightweight probabilistic framework for non-color texture image retrieval, whose main purpose was to reduce the number of operators in calculation of the similarities between the query image and images in the database. One of the innovations of this article is deriving the closed from of KLD among several statistical models such as Generalize Gaussian(GG), Weibull and Rayleigh. In 2011, and following their two previous studies, Kwitt et al. \cite{Kwitt2011} adopted a new approach based on Bayesian frameworks instead of using KLD in the similarity investigation step, and proposed a color texture image retrieval method using the Gaussian Copula, and t models along with several other statistical models. In \cite{Maliani}, Generalized Gamma Distribution and Gaussian Copula function are used for modeling the wavelet domain and presenting a framework for color image retrieval. Maliani has also used KLD as the similarity criterion for his framework. Lasmar et al. \cite{Lasmar}, modeled the wavelet coefficients taking into account the local neighborhood for each sub-band. They used Gaussian Copula function along with GG and Weibull marginal distributions for non-color texture image retrieval. In \cite{Ves}, Generalized Gamma distribution is used for modeling real and imaginary coefficients of the DT-CWT transform. Then the acquired model is used for color texture image retrieval. One of the first researches to model gabor wavelet coefficients for texture image retrieval non-sensitive to rotation, was \cite{Li2015}. In addition, Li \cite{Li2015} used circularly symmetric gabor wavelet to create a platform for retrieval of texture images non-sensitive to image rotation. In 2017, following their previous research, Li et al. \cite{Li2017} introduced three types of dependencies between Gabor Wavelet coefficients with the aid of a model based on Copula and Weibull marginal distribution for color texture image retrieval. In 2018, Karine et al. \cite{Karine} modeled DT-CWT coefficients using Gaussian Copula distribution for retrieving stereo images. In addition to \cite{Karine}, Ghodhbani et al. \cite{Ghodhbani} modeled colored stereo images in the wavelet domain and used the KLD criterion to investigate the level of similarity using the GG model and Copula function. Yang et al. \cite{Yang}, modeled the coefficients of the non-subsampled contourlet transform using the Weibull distribution to design a framework for retrieving grayscale images. Yang has also used KLD as a similarity criterion in his framework. In \cite{Li2019}, continuing his research from 2017, Li has modeled multiple wavelet coefficients using the Marginal Distribution Covariance Model (MDCM). In the MDCM model, the data are first mapped to the Cumulative Distribution Function (CDF) space, and then the covariance model is created in this space. Multiple wavelets in \cite{Li2019} refers to the coefficients from DT-CWT, GWT, and orthogonal wavelet transforms (OWT). \\
\indent In this study, by continuing research on the characteristics of the Shearlet transform (the non-subsampled version) introduced below, we will use Shearlet transform to attain our descriptors. In this study, we proposed a joint modeling of non-subsampled Shearlet coefficients for design of a color texture image retrieval framework using Gaussian Copula function. \\
\indent The contributions of our work are as follows:
\begin{itemize}
 \item Using non-Subsample Shearlet transform to propose a color texture image retrieval framework. The Shearlet domain is used for the first time in the literature for providing a color texture image retrieval framework. 
 \item Presenting a model based on Copula in order to model the dependencies among the non-subsampled Shearlet coefficients. The proposed model is capable of marginal modeling with the help of a variety of non-Gaussian distributions and could be used in various applications for Shearlet coefficients modeling.
 \item Calculating closed form KLD among T Location-Scale (TLS) functions for use in the second step of retrieval, i.e. examining the amount of similarity using the JD criterion (the symmetric version of KLD) which uses KLD. This results in the increase of the retrieval speed of the proposed framework. 
\end{itemize}
\indent The rest of this paper is organized as follows. In the section 2, Shearlet transform and the concept of Copula are introduced. In section 3, first marginal modeling of the Shearlet sub-bands is discussed, and then joint modeling of the Shearlet coefficients with the aid of Copula function is performed, and finally, the color texture image retrieval framework using the joint model is described. A comparison between the proposed framework and the state-of-the-art methods is provided in section 4. Finally, section 5 concludes the paper.
 
\section{BASIC CONCEPTS}
\indent In this section, we first introduce the Shearlet transform and its non-subsampled version, and then introduce the Copula concept and its various functions.\\
\subsection{The Shearlet Transform}
\indent Signal representation in frequency domains has gained a lot of attention from the researchers in recent decades. One of the famous transforms of this field, is wavelet transform. Since wavelet transform has had weaknesses in providing directional features, various transforms have been introduced to improve this weakness, including contourlet transform, curvelet transform, ridgelet transform and shearlet transform.\\
\indent Shearlet transform was first introduced by Lim et al. \cite{Easley} in 2008. This transform is a multi-scale and multi-directional transform that in addition to resolving the problem of displaying multi-directional features, is considered a non-isotropic version of wavelet transform. The non-isotropic property allows for obtaining additional information regarding the geometry of singular points such as the edges and discontinuous points, which are usually seen in multi-dimensional data such as images. Thus, Shearlet transform is a non-isotropic version of the wavelet transform which is capable of extracting details of directional features as well. Continuous shearlet transform of function $f$ in $2D$ space is defined as Eq.~\ref{eq01}: 

\begin{eqnarray}
\label{eq01}
S{H_\psi }f(a,s,t) = \left\langle {f,{\psi _{a,s,t}}} \right\rangle
\end{eqnarray}

where $\left\langle . \right\rangle$ denotes the inner product, $\psi$ is the generating function, $a > 0$ represents the scale parameter, $s \in R$ is the shear parameter, $t \in R^2$ represents the translation parameter, and $\psi _{a,s,t}$  is the basic shearlet function calculated from Eq.~\ref{eq02}:

\begin{eqnarray}
\label{eq02}
{\psi _{a,s,t}}\left( x \right) = {\left| {\det {M_{a,s}}} \right|^{\frac{{ - 1}}{2}}}\psi \left( {M_{a,s}^{ - 1}\left( {x - t} \right)} \right)
\end{eqnarray}

where ${M_{a,s}} = \left[ {\begin{array}{*{20}{c}}
a&{ - \sqrt a s}\\
0&{\sqrt a }
\end{array}} \right]$ Matrix $M_{a,s}$ can also be factorized as Eq.~\ref{eq03}:

\begin{eqnarray}
\label{eq03}
{M_{a,s}} = {B_s}{A_a}
\end{eqnarray}

where
${B_s} = \left[ {\begin{array}{*{20}{c}}
1&{ - s}\\
0&1
\end{array}} \right]$ is the shear matrix and 
${A_a} = \left[ {\begin{array}{*{20}{c}}
a&0\\
0&{\sqrt a }
\end{array}} \right]$ denotes the anisotropic dilation matrix. 

\begin{figure*}[t!]
\begin{center}
\includegraphics[scale=.85]{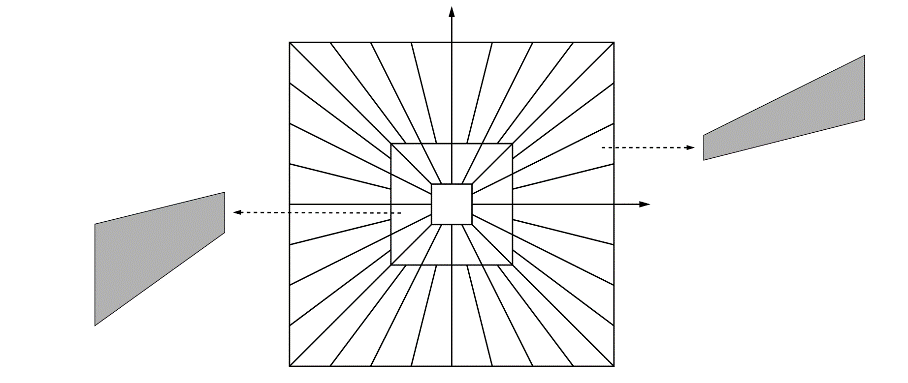}
\end{center}
\caption{\label{fig1} Support for ${\psi _{a,s,t}}$ Shearlets in the frequency domain in shape of trapezoidal pieces.}
\end{figure*}

\indent Each component of ${\psi _{a,s,t}}$ benefits from a frequency support in the form of a trapezoidal pair symmetrical relative to the center which are positioned in different scales $a$ , aligned with the slope $s$ and in location $t$. Fig.~\ref{fig1} demonstrates the frequency support of the shearlet components of ${\psi _{a,s,t}}$ in different scales, location and slope with a trapezoidal form. 

\indent Continuous shearlet transform could be discretized by sampling its scale, shear, and translate \cite{Lim}. The shift invariant version of this transform, is the non-subsample shearlet transform (NSST), which is obtained from the combination of the non-subsampled Laplace pyramid transform and shear filters. The difference between NSST and shearlet transform is in the elimination of high and low samplings. NSST transform is a multi-scale, multi-directional shift invariant transform. Thus, using NSST can help in finding appropriate features for designing an image retrieval system. We have used this transform in the current study to extract features from images.

\subsection{Copula Model}
\indent Copula model, is applicable as a powerful tool in multivariate modeling for a variety of fields including economy, medicine where the multivariate dependencies are of importance. The Copula $C$ is a joint CDF defined in a d-dimensional unit cube ${\left[ {0,1} \right]^d}$ , where every marginal distribution is located in $\left[ {0,1} \right]$ . In fact, the function of Copula is to connect marginal distributions to provide a joint distribution \cite{Jaworski}. \\
\indent The fact that Copula could be useful in reflecting multivariate distribution along with marginal distributions, comes from Sklar’s Theorem. Assume there exists a random d-dimensional vector $\vec X = {\left( {{x_1},...,{x_d}} \right)^t}$ along with its continuous marginal CDFs ${F_1},...,{F_d}$. According to Sklar’s theorem, there exists a unique Copula $C$ that satisfies Eq.~\ref{eq04}.

\begin{eqnarray}
\label{eq04}
F\left( {{x_1},...,{x_d}} \right) = C\left( {{F_1}\left( {{x_1}} \right),...,{F_d}\left( {{x_d}} \right)} \right){\rm{        }}\forall x = \left( {{x_1},...,{x_d}} \right) \in {\Re ^d}
\end{eqnarray}

\indent On the contrary, if $C$ is a Copula function, with ${F_1},...,{F_d}$ denoting its CDF functions, a function $C\left( {{F_1}\left( {{x_1}} \right),...,{F_d}\left( {{x_d}} \right)} \right)$ can be defined which denotes the joint CDF of the vector $\vec X$ along with the marginal CDF functions ${F_1},...,{F_d}$. \\
\indent Now if the function $C$ is continuous and derivable, the Copula density is calculated from Eq.~\ref{eq05} as follows:

\begin{eqnarray}
\label{eq05}
c\left( {{u_1},...,{u_d}} \right) = \frac{{{\partial ^d}C\left( {{u_1},...,{u_d}} \right)}}{{\partial {u_1}...\partial {u_d}}}
\end{eqnarray}

\begin{figure*}[t!]
\begin{center}
\includegraphics[scale=.85]{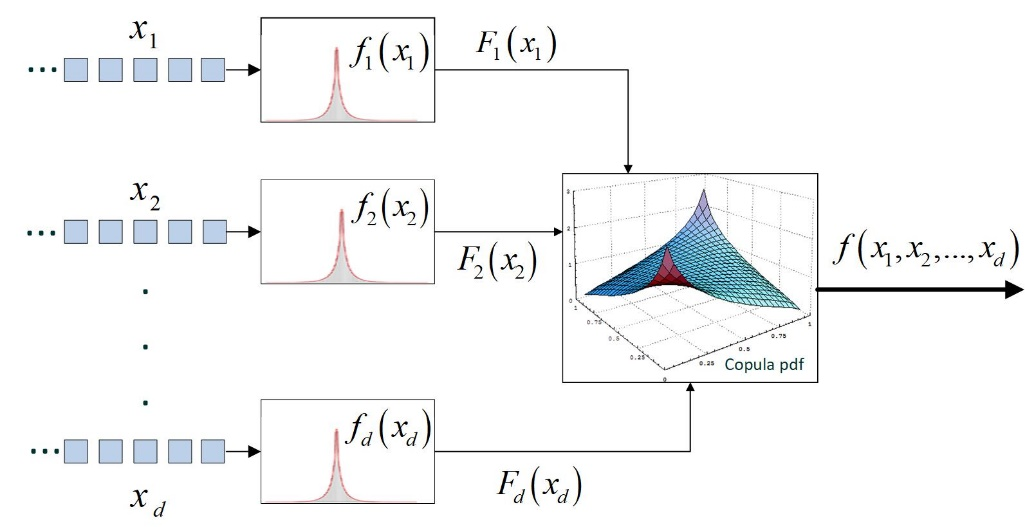}
\end{center}
\caption{\label{fig2} The flowchart for calculating joint distribution of vector $x$ using Copula (Eq.~\ref{eq06}), and marginal models.}
\end{figure*}

\indent In this case, the joint probability density function (PDF) for the vector $\vec X$ is calculated from Eq.~\ref{eq06}:

\begin{eqnarray}
\label{eq06}
f\left( {{x_1},...,{x_d}} \right) = c\left( {{F_1}\left( {{x_1}} \right),...,{F_d}\left( {{x_d}} \right)} \right)\prod\limits_{i = 1}^d {{f_i}\left( {{x_i}} \right)}
\end{eqnarray}

\indent where $f_i \; \forall \: i = 1,...,d$, denote the marginal PDFs of the variables. Therefore, according to Eq.~\ref{eq06}, the joint multivariate PDF, is calculated by considering just the marginal PDF functions of the variables $x_i$ and the Copula density \cite{Jaworski}. Based on Eq.~\ref{eq06}, the joint multivariate distributions could be derived as seen in Fig.~\ref{fig2}.

\begin{figure*}[t!]
\begin{center}
\includegraphics[scale=.65]{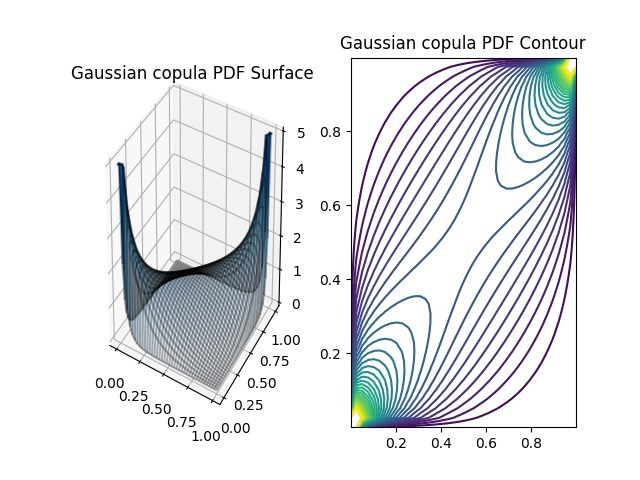}
\end{center}
\caption{\label{fig3} The Gaussian Copula function in two- and three-dimensional spaces.}
\end{figure*}

\indent Use of Eq.~\ref{eq06} to calculate the joint distribution has the advantage that modeling the marginal distributions of the variables becomes separate from modeling the dependency distribution among the variables. Using this method enables us to use various distributions in modeling the marginal distribution and dependency distribution among the variables. The family of Copula functions is generally divided into two main categories of elliptic Copulas and Archimedean Copulas, that are different in their types of dependency reflection. Gaussian Copula and Student’s t Copula are among the elliptic Copula functions, while Clayton, Gumbel, and Frank are some of the well-known Archimedean Copulas. In the family of elliptic copulas, the Copula known as Gaussian has attracted the most attention among the researchers. The reason for this is its easy implementation, because it models the dependency structure based on correlation coefficients. Furthermore, its related hyperparameters are easily estimated using maximum likelihood-based estimators.\\ 
\indent Rewriting the multivariate Gaussian distribution and considering Eq.~\ref{eq06}, the Gaussian Copula density would be defined by Eq.~\ref{eq07}:

\begin{eqnarray}
\label{eq07}
c\left( {{u_1},{u_2},...,{u_d}} \right) = \frac{1}{{{{\left| \Sigma  \right|}^{{1 \mathord{\left/
 {\vphantom {1 2}} \right.
 \kern-\nulldelimiterspace} 2}}}}}\exp \frac{{ - {{\vec y}^t}\left( {{\Sigma ^{ - 1}} - I} \right)\vec y}}{2}
\end{eqnarray}

\indent where ${\vec y^t} = \left( {{y_1},...,{y_d}} \right)$ denotes the transpose of vector $\vec y$ that elements are calculated as ${y_i} = {\phi ^{ - 1}}\left( {{u_i}} \right)$ , $\phi $  denotes the CDF of the standard Gaussian distribution. Matrix $I$ is a d-dimensional identity matrix and $\Sigma$ denotes the correlation matrix. Fig.~\ref{fig3} illustrates an example of the Gaussian Copula function in two- and three-dimensional spaces.

\section{Modeling of NSST Coefficients and the Proposed Retrieval Framework}
\indent In this section, we first discuss marginal modeling of NSST coefficients. Then, different types of neighboring among NSST sub-bands and coefficients are defined, and the dependencies between various models of neighboring are investigated. We will then investigate joint modeling of coefficients using Copula functions. Finally, we will introduce our proposed retrieval framework based on joint modeling using Copula function.\\

\subsection{Modeling NSST coefficients using Copula function}
\indent Many previous articles in the literature have investigated marginal modeling of frequency domain coefficients without considering their dependencies \cite{Etemad2018,Etemad2016,Rabizadeh,Sadreazami}. Considering the dependencies among the coefficients in multi-directional and multi-scale frequency transforms could improve the modeling of coefficients in these transforms. In this paper, we have focused on joint modeling of NSST coefficients using Copula function. \\
As seen from Eq.~\ref{eq06}, joint modeling of NSST coefficients using the Copula function includes two steps: 1) marginal modeling of NSST coefficients, 2) joint modeling of NSST coefficients using the appropriate Copula function. 

\subsubsection{Marginal Modeling of NSST Coefficients}
\begin{figure*}[t!]
\begin{center}
\includegraphics[scale=.70]{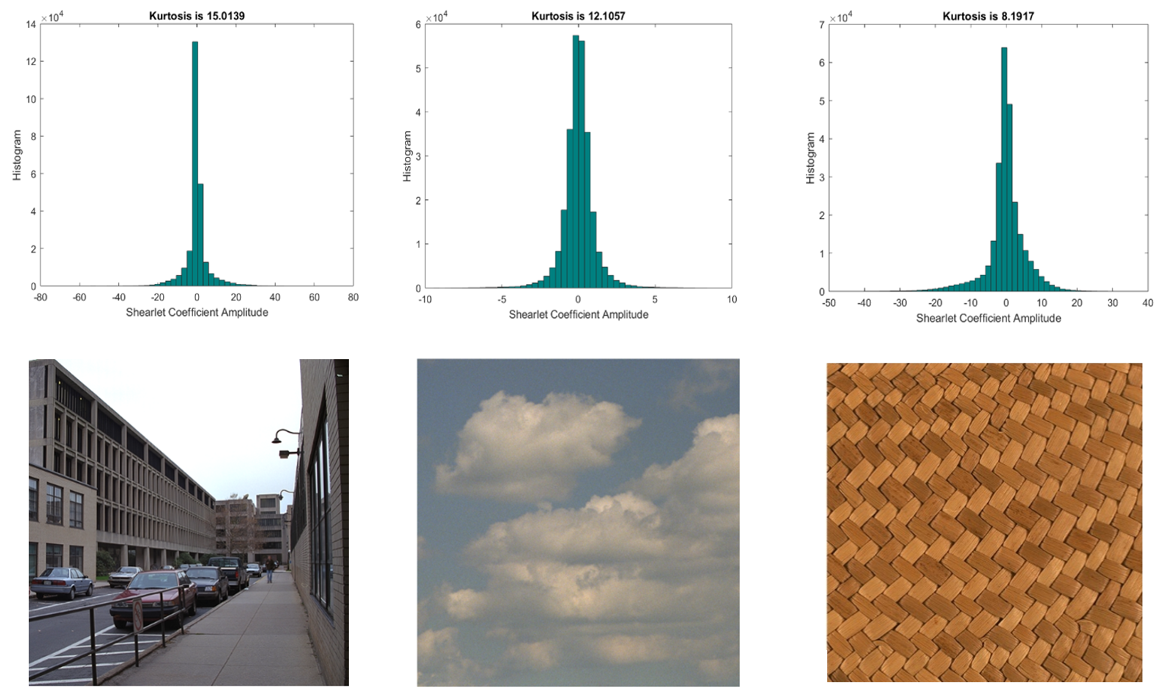}
\end{center}
\caption{\label{fig4} Histogram of NSST coefficients corresponding to the red channel of the three texture images \cite{Vistex} extracted in scale of 2 and with 4 directions.}
\end{figure*}

\indent It is shown in \cite{Wang} that marginal distribution of shearlet coefficients for different scales and directions in medical images contains sharp peaks around zero and has a non-gaussian distribution with heavy tails. Fig.~\ref{fig4} shows three images from the dataset \cite{Vistex}, along with the histogram for NSST coefficients with parameters of scale 2 and direction 4. As can be seen from Fig.~\ref{fig4}, histogram of the coefficients contains sharp peaks around zero along with a non-Gaussian distribution with heavy tails. Moreover, the kurtosis of the histograms is much higher than 3, the kurtosis of the gaussian distribution, which is an indication of non-Gaussian nature of coefficients distribution. It should be noted that similar results were obtained for all images and their corresponding sub-bands. Thus, for marginal modeling of NSST coefficients, non-Gaussian distributions, such as T location-scale (TLS), and Generalized Gaussian (GG) distributions should be exploited.
 
\indent In this article, we have used GG and TLS distributions for marginal modeling of the coefficients. 
The GG distribution function of random variable x is defined by Eq.~\ref{eq08}:

\begin{eqnarray}
\label{eq08}
f\left( {x;\alpha ,\beta ,\mu } \right) = \frac{\beta }{{2\alpha \Gamma \left( {\frac{1}{\beta }} \right)}}\exp \left( { - {{\left( {\frac{{\left| {x - \mu } \right|}}{\alpha }} \right)}^\beta }} \right){\rm{   }}x \in R
\end{eqnarray}

\indent where $\mu \in R $ is the location parameter, $\alpha > 0$ is the scaling factor, 
$\beta  > 0$  represents the shape parameter and 
$\Gamma \left( z \right) = \int_0^\infty  {{t^{z - 1}}{e^{ - t}}dt} $  is the gamma function. The three parameters $\alpha$, $\beta$, and $\mu$ are estimated using the maximum likelihood (ML) method \cite{Do}.

\indent The TLS distribution function of the random variable $x$ is defined by Eq.~\ref{eq09} as follows:
\begin{eqnarray}
\label{eq09}
f(x;\mu ,\sigma ,v) = \frac{{\Gamma (\frac{{v + 1}}{2})}}{{\sigma \sqrt {v\pi } \Gamma (\frac{v}{2})}}{(1 + \frac{1}{v}{(\frac{{x - \mu }}{\sigma })^2})^{ - (\frac{{v + 1}}{2})}}
\end{eqnarray}

\indent where in Eq.~\ref{eq09}, $\mu \in R$ denotes the location parameter,
$\alpha > 0$ is the scaling factor, and $\nu$ is the shape parameter (degree of freedom) of the TLS distribution. 

\subsubsection{Joint Modeling of NSST Coefficients using Copula Function}
\begin{figure*}[t!]
\begin{center}
\includegraphics[scale=.75]{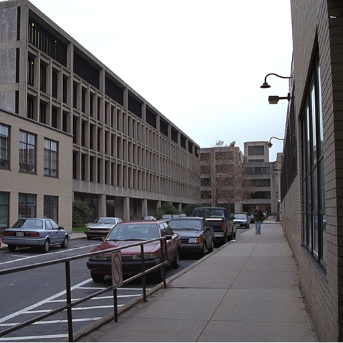}
\subcaption{}
\includegraphics[scale=.75]{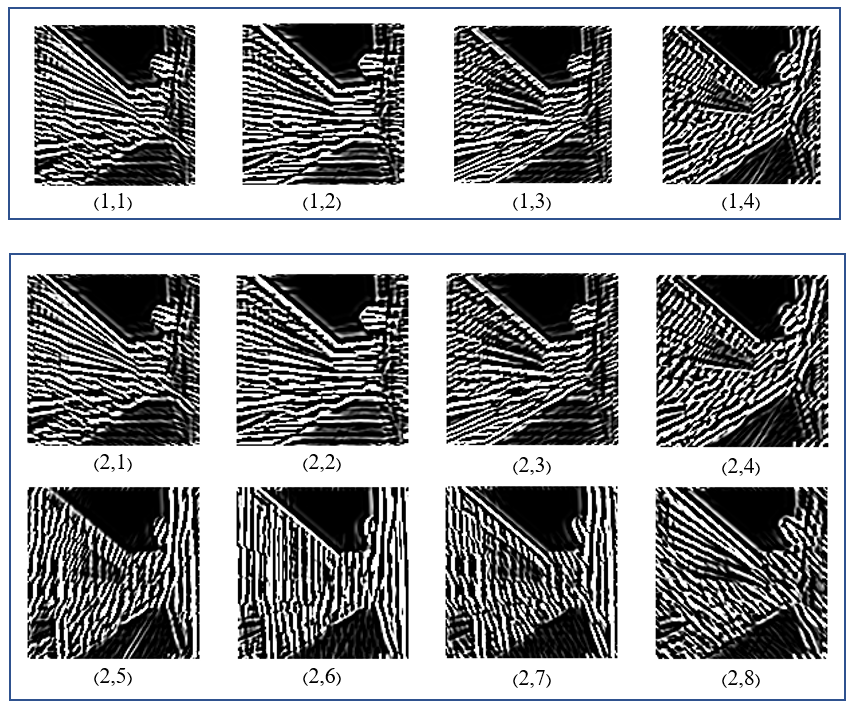}
\subcaption{}
\end{center}
\caption{\label{fig5} Images of NSST coefficients corresponding to the red channel of a color image from the dataset \cite{Vistex}; a) original image, b) first- and second- scale sub-bands with 4 and 8 different directions, respectively.}
\end{figure*}

\indent It is well known that there exist different color standards for color images, one of which is the RGB standard, wherein each color image consists of three channels, namely red, green and blue. Fig.~\ref{fig5} demonstrates sub-bands of non-subsampled shearlet transform with two scales and twelve directions (4 directions in the first scale and 8 in the second) for the red channel of color images. The numbers in the double tuple $\left( {s,d} \right)$ represent the number of the decomposition scale $s$ and direction $d$, respectively. Every pixel in the sub-band is equal to its corresponding NSST coefficient in the defined scales and directions.

\begin{figure*}[t!]
\begin{center}
\includegraphics[scale=.75]{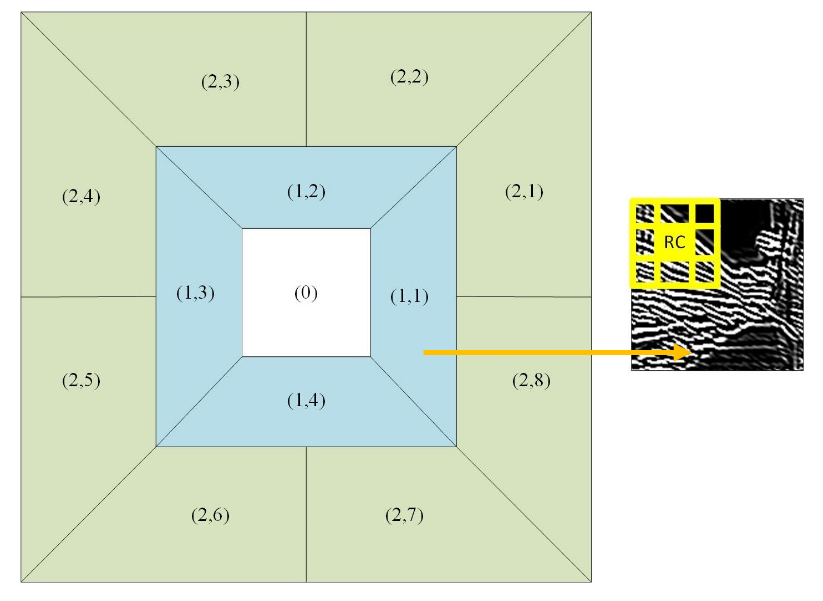}
\end{center}
\caption{\label{fig6} Schematics of non-subsampled shearlet transform in three different scales.}
\end{figure*}

\indent In general, four types of neighborhood can be defined for non-subsampled shearlet coefficients in color images: 1) inter-scale, 2) inter-direction, 3) inter-channel, and 4) local intra-scale.\\
\indent For better understanding of these different types of neighborhoods, see Fig.~\ref{fig6}. Fig.~\ref{fig6} presents the schematics for non-subsampled shearlet transform in three scales and for 12 different directions. The scale (0) represents the low-frequency coefficients. Each of the trapezoids inside the figure represent a sub-band inside the decomposition scale for a specific direction which include the NSST coefficients. Examples of different types of neighborhoods shown in Fig.~\ref{fig6} include: 1) inter-scale neighborhood: neighborhood between coefficients of the sub-band (1,x), and coefficients of the (2,x) sub-bands, 2) inter-direction neighborhood: neighborhood among the coefficients of (1,x) sub-bands, 3)inter-channel neighboring for colored channels: neighborhood among the coefficients of sub-band (1,x) derived from three different color channels, and 4)local intra-scale neighborhood: reference coefficient (RC) neighboring inside a sub-band such as (1,1) in the form of proximity of sub-band pixels (see Fig.~\ref{fig6}).

\begin{figure*}
\begin{center}
    \begin{subfigure}[t]{0.45\textwidth}
        \centering
        \includegraphics[scale=.52]{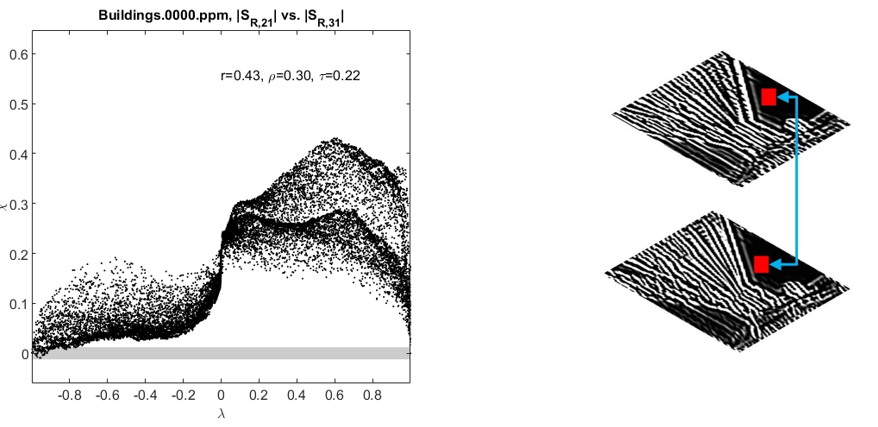}
        \caption{}
    \end{subfigure}\
    
  \vspace*{6pt}%
   
    \begin{subfigure}[t]{0.45\textwidth}
        \centering
        \includegraphics[scale=.45]{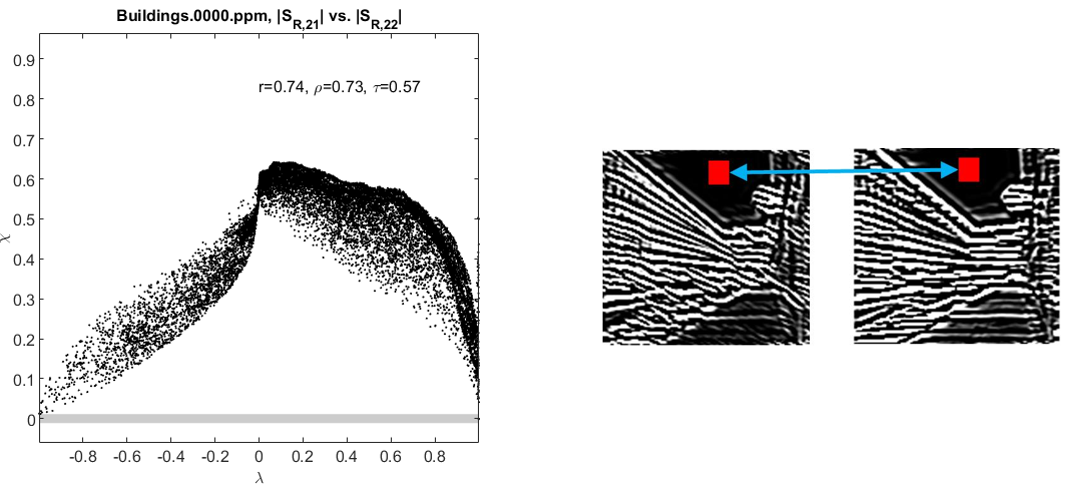}
        \caption{}
    \end{subfigure}
    
  \vspace*{6pt}%
   
    \begin{subfigure}[t]{0.45\textwidth}
        \centering
        \includegraphics[scale=.45]{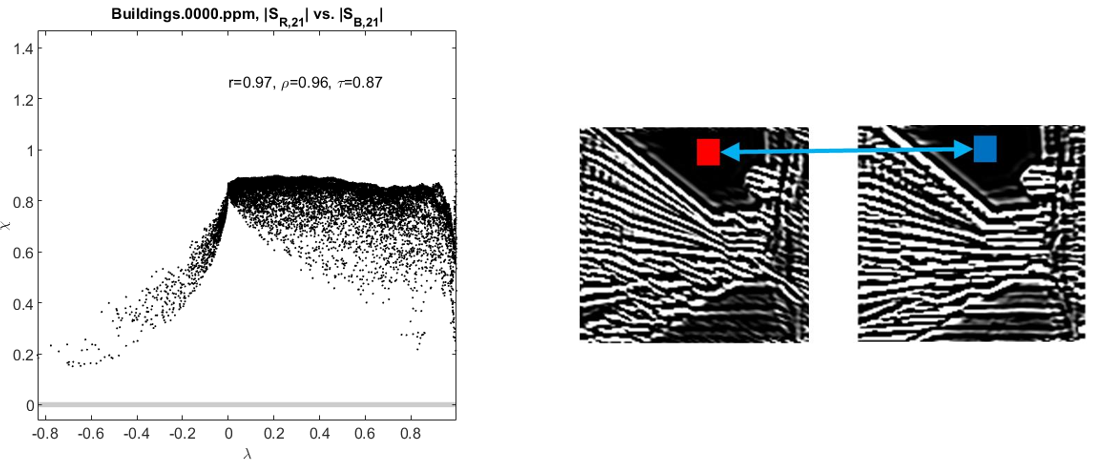}
        \caption{}
    \end{subfigure}
    
  \vspace*{6pt}%
  
    \begin{subfigure}[t]{0.45\textwidth}
        \centering
        \includegraphics[scale=.45]{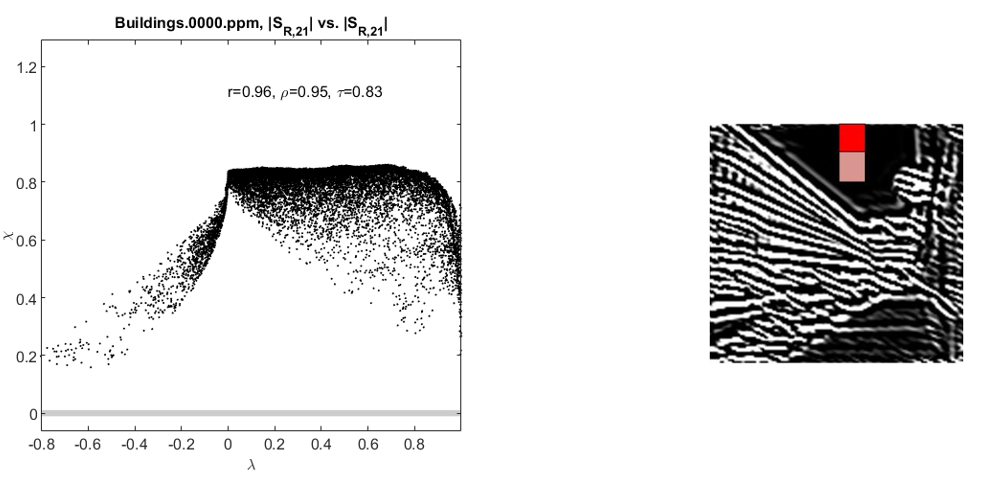}
        \caption{}
    \end{subfigure}
    ~
\caption{\label{fig7}Chi-plot of the NSST coefficients pair in form of different neighborhoods a) inter-scale, b) inter-direction, c) inter-channel, and d) intra-scale.
${S_{X,YZ}}$  denotes the sub-band of $S$ in the color channel $X$ with shear scale $Y$ in direction $Z$.}
\end{center}
\end{figure*}

\indent We used Chi-plot to examine the dependencies among NSST coefficients in the defined neighborhoods. Chi-plot was first introduced by Fischer et al. \cite{Fisher} to recognize dependencies among various variables. Fig.~\ref{fig7} shows the chi-plot for the pair of NSST coefficients in form of different neighborhoods for the original image shown in Fig.~\ref{fig5}. We have used the settings defined in \cite{Fisher} for plotting this figure. The grayscale stripe inside this figure is called the tolerance band. The degree of deviation of the plotted tolerance band points, demonstrate the dependencies among the coefficients. As seen from Fig.~\ref{fig7}, the degree of deviation in the inter-direction, intra-scale and inter-channel neighborhoods are higher than the inter-scale. In addition, the values of correlation coefficient, spearman coefficient and kendall’s tau in Fig.~\ref{fig7} confirm this issue. Of course, this does not mean that no dependencies exist among the NSST coefficients in the inter-scale case, but is an indication of lower dependencies in this neighboring case. It should be noted that similar results were obtained for other images in the datasets.

\begin{figure*}[t!]
\begin{center}
\includegraphics[scale=.55]{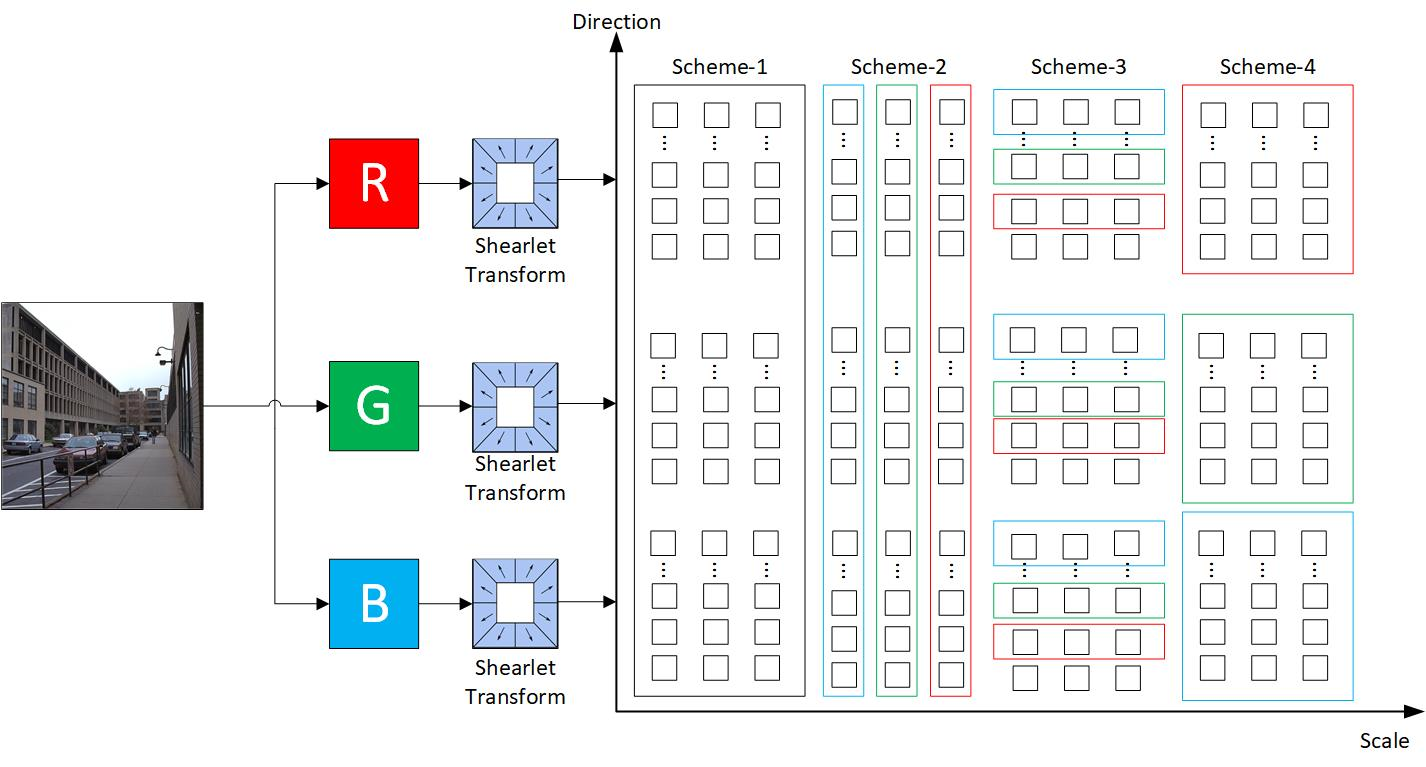}
\end{center}
\caption{\label{fig8} Different types of dependencies among sub-bands of NSST for color images. In each scheme, three NSST transform scales are applied to each color channel. Each small square represents a NSST sub-band. The sub-bands inside each rectangle are modeled along each other.}
\end{figure*}

\indent Four different schemes could be defined for joint modeling of NSST coefficients of color images based on the defined neighborhoods (except for intra-scale neighborhood) among NSST coefficients. Fig.~\ref{fig8} shows these four schematics. In defining the first scheme, all inter-scale, inter-direction, and inter-channel dependencies are considered. In the second diagram, inter-direction and inter-channel dependencies are considered. The third scheme considers inter-scale and inter-channel dependencies, while the fourth scheme considers inter-direction and inter-scale dependencies. Computational cost and modeling precision are considered two decisive parameters in choosing the final scheme for modeling NSST coefficients. Naturally, the first scheme has the highest computational cost and retrieval precision among all the other options, but it remains to be seen to what extent this computational cost will cause problems in practice. In this article, after studying the different schemes in section four, we have used the first one for joint modeling of NSST coefficients.

\indent As stated in the previous section, the Copula function contains various families that are used for different applications. Since we are searching for a color texture image retrieval framework in this study, this issue should be considered when choosing the Copula function. The Gaussian Copula function has gained a lot of attention among the researchers due to ease of parameter estimation and existence of closed form KLD in texture image retrieval \cite{Lasmar,Li2017}. Thus, we will also use the Gaussian Copula function as the Copula pdf. \\
\indent Based on Eq.~\ref{eq06}, the joint distribution function of NSST coefficients depends on the type of Copula function and marginal distributions are generated in various forms. In Eq.~\ref{eq10}, 
${f_{GC - GG}}$ denotes joint distribution of the coefficients of 
$\overrightarrow X  = \left( {{x_1},{x_2},...,{x_d}} \right)$, assuming GG and Gaussian Copula distributions:

\begin{eqnarray}
\label{eq10}
\begin{array}{l}
{f_{GC - GG}}\left( {\overrightarrow X ;\theta } \right) = \frac{1}{{{{\left| \Sigma  \right|}^{1/2}}}}\exp \frac{{ - {{\vec y}^t}\left( {{\Sigma ^{ - 1}} - I} \right)\vec y}}{2} \times {\left( {\frac{\beta }{{2\alpha \Gamma \left( {\frac{1}{\beta }} \right)}}} \right)^d}\exp \left( { - \sum\limits_{i = 1}^d {{{\left( {\frac{{\left| {{x_i} - \mu } \right|}}{\alpha }} \right)}^\beta }} } \right)
\end{array}
\end{eqnarray}

\indent where 
$\theta  = \left( {\Sigma ,\alpha ,\beta ,\mu } \right)$  is the hyperparameter of joint distribution of the coefficients of $X$ and 
${y_i} = {\phi ^{ - 1}}\left( {F\left( {{x_i};\alpha ,\beta ,\mu } \right)} \right)$. Considering the TLS distribution instead of the GG distribution, a new distribution obtained by the name of 
${f_{GC - TLS}}$ as Eq.~\ref{eq11}:

\begin{eqnarray}
\label{eq11}
\begin{array}{l}
{f_{GC - TLS}}(\overrightarrow X ;\theta ) = \frac{1}{{{{\left| \Sigma  \right|}^{{1 \mathord{\left/
 {\vphantom {1 2}} \right. \kern-\nulldelimiterspace} 2}}}}}\exp \frac{{ - {y^T}\left( {{\Sigma ^{ - 1}} - I} \right)y}}{2} \times \prod\limits_{i = 1}^d {\frac{{\Gamma (\frac{{v + 1}}{2})}}{{\sigma \sqrt {v\pi } \Gamma (\frac{v}{2})}}{{(1 + \frac{1}{v}{{(\frac{{{x_i} - \mu }}{\sigma })}^2})}^{ - (\frac{{v + 1}}{2})}}}
\end{array}
\end{eqnarray}

\indent where 
$\theta  = \left( {\Sigma ,\mu ,\alpha ,v} \right)$  is the hyperparameter of joint distribution of the coefficients of $X$ and 
${y_i} = {\phi ^{ - 1}}\left( {F\left( {{x_i};\mu ,\alpha ,v} \right)} \right)$ .  

\begin{figure*}
\begin{center}
    \begin{subfigure}[t]{0.45\textwidth}
        \centering
        \includegraphics[scale=.45]{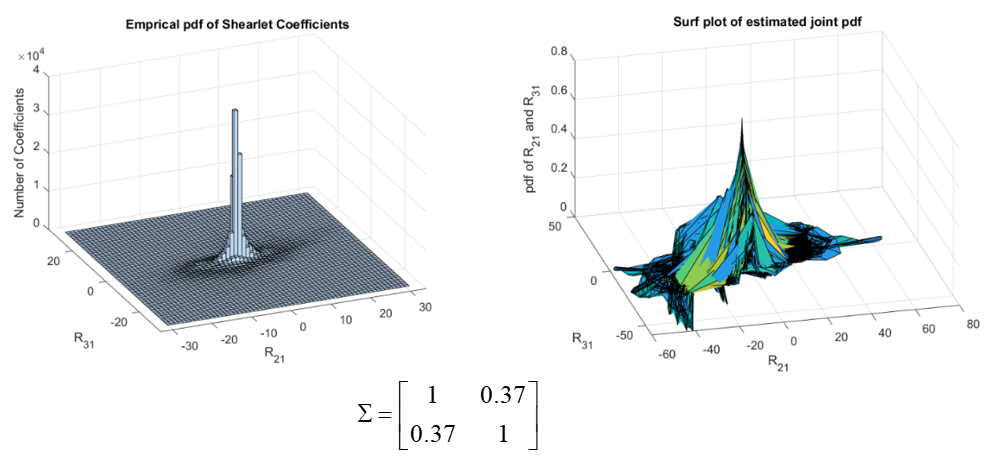}
        \caption{}
    \end{subfigure}\
    
  \vspace*{6pt}%
   
    \begin{subfigure}[t]{0.45\textwidth}
        \centering
        \includegraphics[scale=.45]{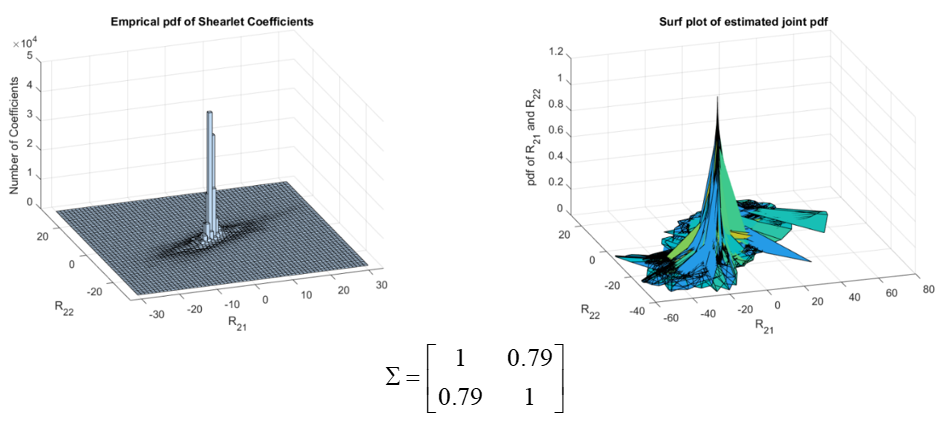}
        \caption{}
    \end{subfigure}
    
  \vspace*{6pt}%
   
    \begin{subfigure}[t]{0.45\textwidth}
        \centering
        \includegraphics[scale=.45]{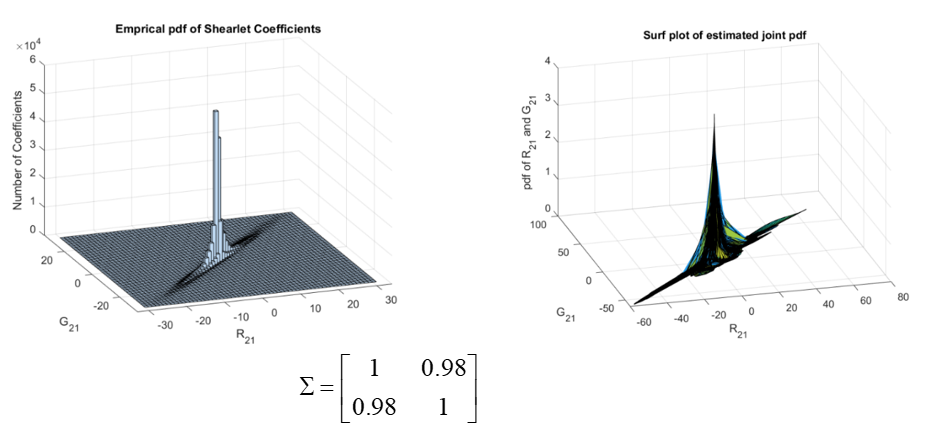}
        \caption{}
    \end{subfigure}
    
  \vspace*{6pt}%
  
    \begin{subfigure}[t]{0.45\textwidth}
        \centering
        \includegraphics[scale=.45]{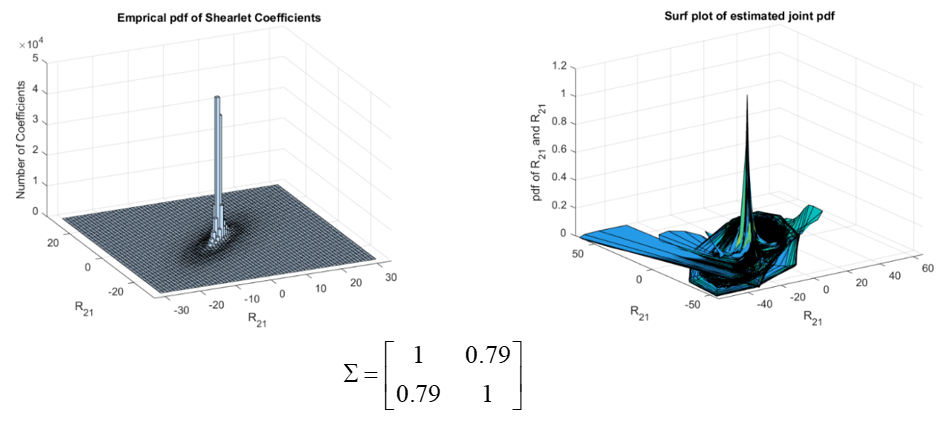}
        \caption{}
    \end{subfigure}
    ~
\caption{\label{fig9} Joint histogram of NSST coefficients in form of a) inter-scale, b) inter-direction, c) inter-channel, and d) intra-scale Neighborhood along with the approximated joint distribution function ${f_{GC - GG}}$. }
\end{center}
\end{figure*}

\indent Fig.~\ref{fig9} shows the joint histogram for different types of neighboring in hypothetical color texture image (image shown in Fig.~\ref{fig5}), and the approximate function
${f_{GC - GG}}$. It can be inferred from Fig.~\ref{fig9} that using Gaussian Copula function with linear dependence (correlation) for modeling NSST coefficients is a right choice because histograms also show high linear dependence in different cases. The value of the Gaussian Copula function parameter $\Sigma$ for each neighborhood is specified in the figure. It should be noted that similar results were obtained for other color texture images as well as ${f_{GC - TLS}}$ function.

\begin{figure*}[t!]
\begin{center}
\includegraphics[scale=.50]{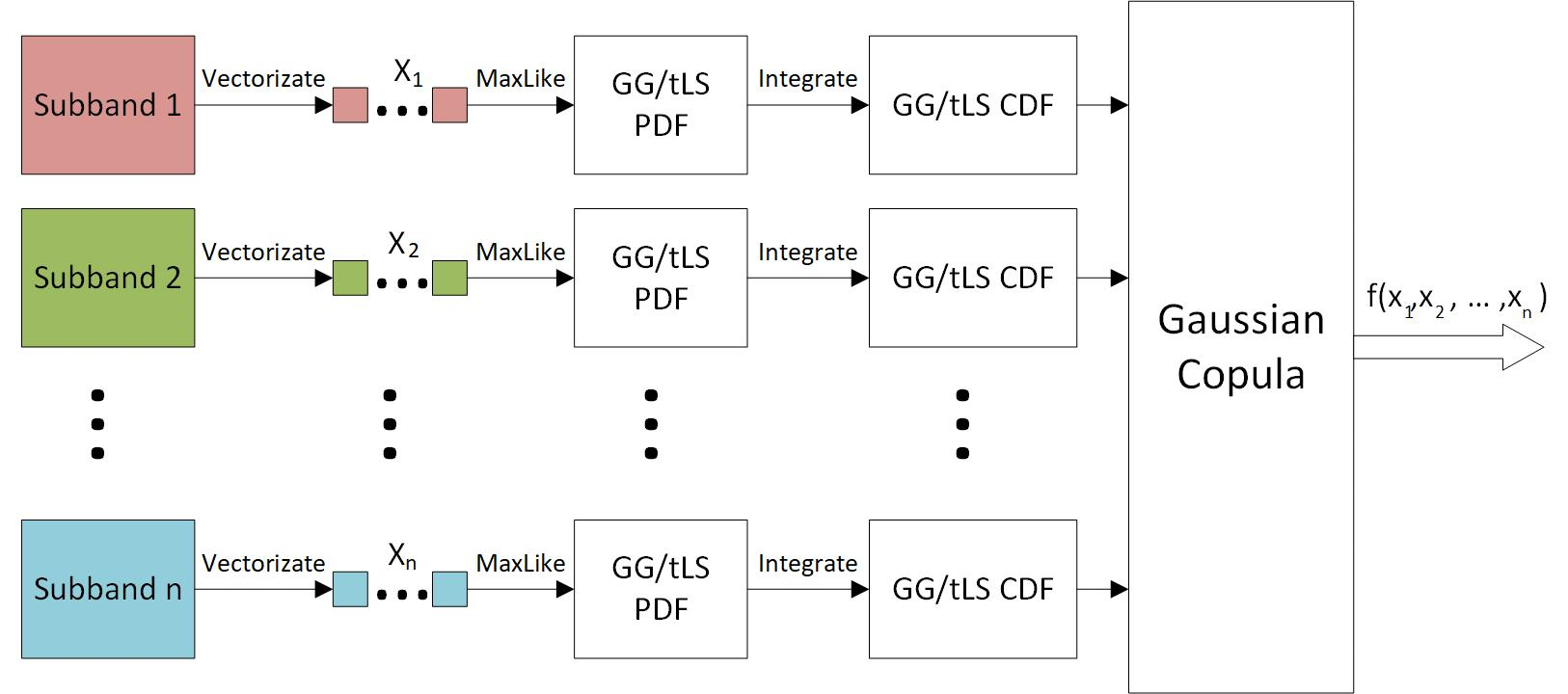}
\end{center}
\caption{\label{fig10} How to build GC-GG and GC-TLS models for modeling NSST sub-bands.}
\end{figure*}

\indent Fig.~\ref{fig10} demonstrates how to calculate Eqs.~\ref{eq10}, and ~\ref{eq11}. In the first step, the sub-bands of color channels are calculated for different directions and on a specific scale (for example the second scale), using NSST transform. Then, each of the sub-bands with dimensions
$(m \times n)$ is transformed into vectors with dimensions 
$\left( {(m \times n),1} \right)$, using vectorization. Next, maximum likelihood method is used to estimate the marginal distribution parameters and their pdfs are calculated. Following that, integrating the pdf function, the CDF is calculated for each of the vectors, and sent to the Gaussian Copula function as an input. Finally, multiplying the Gaussian Copula function by pdf functions of each vector, the joint distribution function of NSST coefficients is calculated. 

\subsection{Color Texture Image Retrieval Framework using GC-GG and GC-TLS Models}
\begin{figure*}[t!]
\begin{center}
\includegraphics[scale=.50]{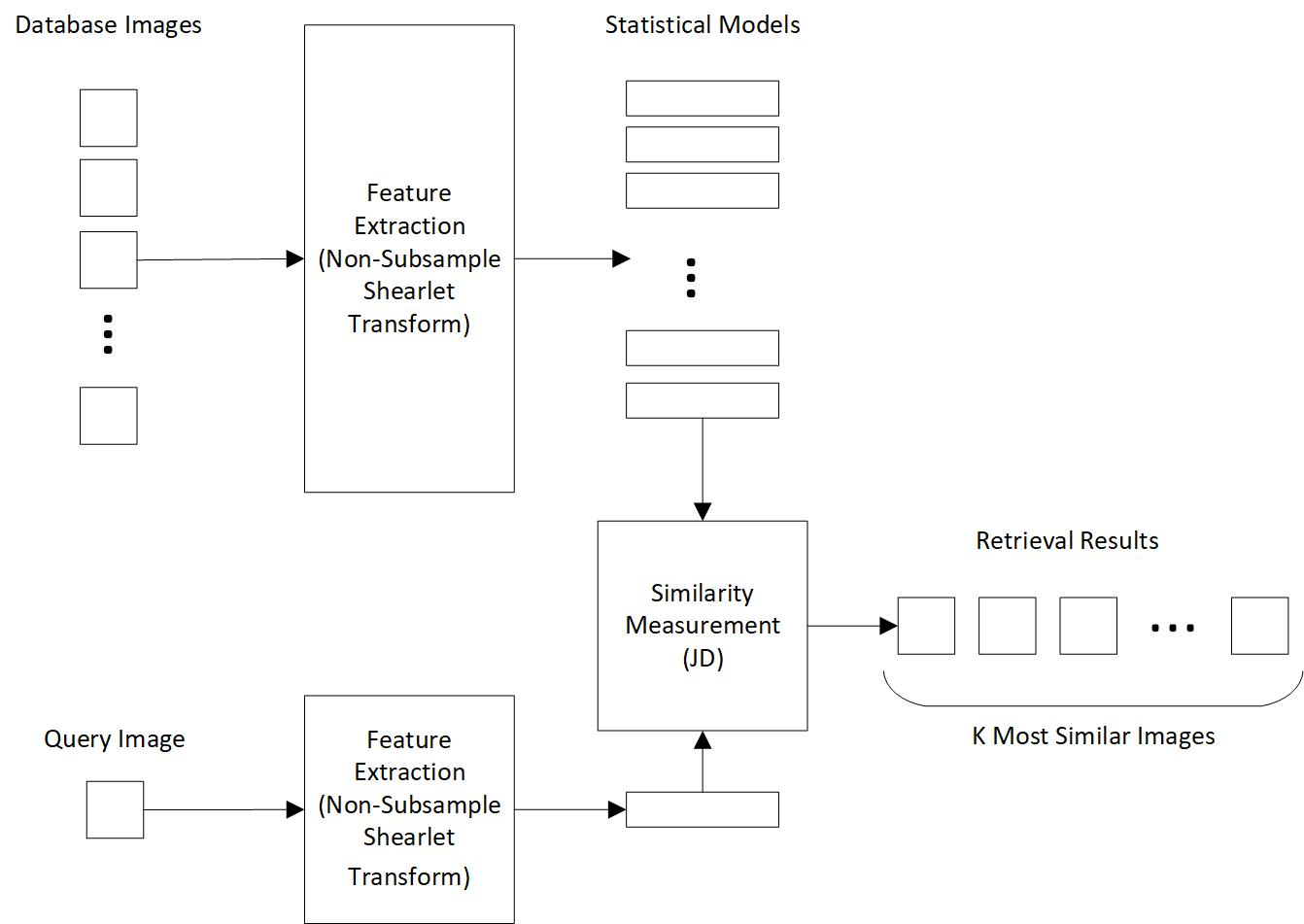}
\end{center}
\caption{\label{fig11} The diagram schematics of image retrieval system using statistical modeling.}
\end{figure*}

\indent As mentioned before, image retrieval includes two main steps of feature extraction (along with modeling) and similarity analysis. Fig.~\ref{fig11}, demonstrates the proposed image retrieval framework with statistical modeling of NSST coefficients. 

\indent As seen from Fig.~\ref{fig11}, we first apply the NSST transform on all the images in the database (the feature extraction step). Then we model NSST coefficients using GC-GG or GC-TLS models. We perform the same steps on the query image. In the second step, JD criterion (the symmetric version of KLD) is used to determine similarities between the query image distribution and images in the database (the similarity analysis step). In general, the KLD between two PDFs, $f$ and $g$ is calculated using Eq.~\ref{eq12}:

\begin{eqnarray}
\label{eq12}
KLD\left( {f\left( {\overrightarrow x ;{\theta ^{db}}} \right)||g\left( {\overrightarrow x ;{\theta ^q}} \right)} \right) = \int {f\left( {\overrightarrow x ;{\theta ^{db}}} \right)} \log \left( {\frac{{f\left( {\overrightarrow x ;{\theta ^{db}}} \right)}}{{g\left( {\overrightarrow x ;{\theta ^q}} \right)}}} \right)dx
\end{eqnarray}

where 
$\theta  = \left( {{\theta ^{db}},{\theta ^q}} \right)$  are the hyper parameters of functions $f$ and $g$, which have been used for modeling the images in the database and the query image, respectively. In this paper, we have first used the symmetric version of KLD, known as Jeffery Divergence (JD), which is defined in Eq.~\ref{eq13}:

\begin{eqnarray}
\label{eq13}
\begin{array}{l}
JD\left( {f\left( {\overrightarrow x ;{\theta ^{db}}} \right)||g\left( {\overrightarrow x ;{\theta ^q}} \right)} \right) = KLD\left( {f\left( {\overrightarrow x ;{\theta ^{db}}} \right)||g\left( {\overrightarrow x ;{\theta ^q}} \right)} \right) + \\
KLD\left( {g\left( {\overrightarrow x ;{\theta ^q}} \right)||f\left( {\overrightarrow x ;{\theta ^{db}}} \right)} \right)
\end{array}
\end{eqnarray}

\indent Finally, $k$ images from the database images with highest similarity to the query image are chosen.\\
\indent The proposed color texture image retrieval framework consists of the following steps:

\begin{enumerate}
\item Applying NSST transform on the color texture image assuming $3 \times S$  decomposition scales and $3 \times D$ directions on the red, green, and blue color channels.
\item Choosing sub-bands based on one of the schemes 1 to 4 in Fig.~\ref{fig8} and vectorization of the chosen sub-bands and their modeling using the GC-GG or GC-TLS models as shown in Fig.~\ref{fig10}.
\item Using GG or TLS distributions and Gaussian Copula function for modeling. 
\item Using maximum likelihood method for estimating the hyperparameters $\theta $  of the GC-GG or GC-TLS models.
\item Using JD as a similarity criterion between GC-GG or GC-TLS models corresponding to the query image and images in the database for retrieving similar texture images.
\end{enumerate}

\indent In the last step of the proposed framework, calculation of KLD among the two GC-GG or GC-TLS models is required. Numerically, the KLD criterion is calculated using the maximum likelihood method based on the Bayesian rule, while in some cases a closed form of this can be calculated. The advantage of calculating the closed form of KLD in comparison to the ML-based method, is its much lower computational cost. Because the closed form just uses model parameters instead of number of the coefficients (equal to pixel dimensions of the image). Thus, in the following we will investigate calculation of the KLD closed form for GC-GG and GC-TLS models.

\indent Assume the model $f$ is used for modeling the database images and model $g$ is used for modeling the query image. Assuming they both use a Copula-based model, we have:

\begin{eqnarray}
\label{eq14}
f\left( {\overrightarrow x ;{\theta ^{db}}} \right) = c\left( {{F_1}\left( {{x_1}} \right),{F_2}\left( {{x_2}} \right),...,{F_d}\left( {{x_d}} \right)} \right)\prod\limits_{i = 1}^d {{f_i}\left( {{x_i}} \right)}
\end{eqnarray}

\begin{eqnarray}
\label{eq15}
g\left( {\overrightarrow x ;{\theta ^q}} \right) = c\left( {{G_1}\left( {{x_1}} \right),{G_2}\left( {{x_2}} \right),...,{G_d}\left( {{x_d}} \right)} \right)\prod\limits_{i = 1}^d {{g_i}\left( {{x_i}} \right)}
\end{eqnarray}

\indent where the hyperparameters ${\theta ^{db}}$ and ${\theta ^q}$ are defined in Eq.~\ref{eq16}, respectively:

\begin{eqnarray}
\label{eq16}
{\theta ^{db}} = \left\{ {\left( {\eta _1^{(db)},\eta _2^{(db)},...,\eta _d^{(db)}} \right),{\Sigma _{db}}} \right\}{\rm{ , }}{\theta ^q} = \left\{ {\left( {\eta _1^{(q)},\eta _2^{(q)},...,\eta _d^{(q)}} \right),{\Sigma _q}} \right\}
\end{eqnarray}

\indent Where assuming the GG distribution of the value ${\eta _x} = \left( {{\alpha _x},{\beta _x},{\mu _x}} \right)$ will be obtained and assuming the TLS distribution, the value ${\eta _x} = \left( {{\mu _x},{\alpha _x},{v_x}} \right)$ is obtained.

\indent The closed form KLD between the two Copula-based models ${f_{db}}$ and ${g_q}$ is as follows:

\begin{eqnarray}
\label{eq17}
KLD\left( {{f_{db}}||{g_q}} \right) = KLD\left( {{c_{db}}||{c_q}} \right) + \sum\limits_{i = 1}^d {KLD\left( {f_i^{db}||g_i^q} \right)}
\end{eqnarray}

\indent The first term in Eq.~\ref{eq17}, calculates KLD among the two Copula pdf functions ($c$) and the second term provides a calculation of the KLD among the two marginal distribution functions ($f$ and $g$). Depending on the type of the joint modeling schematics, the first term in Eq.~\ref{eq17} could be written in various forms. Assume $M = C \times D \times S$ represents the total number of NSST sub-bands, where $D$ is the number of directions extracted, $S$ denotes the number of decomposition scales, and $C=3$ is the number of color channels. Assuming the schemes one to four in Fig.~\ref{fig8}, Eqs.~\ref{eq18} to ~\ref{eq21} could be obtained. Eq.~\ref{eq18} for the first scheme:

\begin{eqnarray}
\label{eq18}
KLD\left( {{f_{db}}||{g_q}} \right) = KLD\left( {c_{db}^j||c_q^j} \right) + \sum\limits_{i = 1}^M {KLD\left( {f_i^{db}||g_i^q} \right)}
\end{eqnarray}

And Eqs.~\ref{eq19}, ~\ref{eq20}, and ~\ref{eq21} for schemes 2, 3, and 4:

\begin{eqnarray}
\label{eq19}
KLD\left( {{f_{db}}||{g_q}} \right) = \sum\limits_{j = 1}^S {KLD\left( {c_{db}^j||c_q^j} \right)}  + \sum\limits_{i = 1}^M {KLD\left( {f_i^{db}||g_i^q} \right)}
\end{eqnarray}

\begin{eqnarray}
\label{eq20}
KLD\left( {{f_{db}}||{g_q}} \right) = \sum\limits_{j = 1}^D {KLD\left( {c_{db}^j||c_q^j} \right)}  + \sum\limits_{i = 1}^M {KLD\left( {f_i^{db}||g_i^q} \right)}
\end{eqnarray}

\begin{eqnarray}
\label{eq21}
KLD\left( {{f_{db}}||{g_q}} \right) = \sum\limits_{j = 1}^3 {KLD\left( {c_{db}^j||c_q^j} \right)}  + \sum\limits_{i = 1}^M {KLD\left( {f_i^{db}||g_i^q} \right)}
\end{eqnarray}

\indent In the remainder of this paper we use the first schematic for modeling. The reason for this choice is the higher precision of the model in modeling the coefficients while considering all dependencies. The equations in the remainder of this paper could be easily rewritten for all other schemes. Now by assuming the first schematics in Fig.~\ref{fig8}, and considering the Gaussian Copula function denoted by $c$, we have:

\begin{eqnarray}
\label{eq22}
\begin{array}{l}
KLD\left( {f\left( {\overrightarrow x ;{\theta ^{db}}} \right)||g\left( {\overrightarrow x ;{\theta ^q}} \right)} \right) = \\ 
\frac{1}{2}\left( {tr\left( {\Sigma _q^{ - 1}{\Sigma _{db}}} \right) + \log \frac{{\left| {{\Sigma _q}} \right|}}{{\left| {{\Sigma _{db}}} \right|}} - M} \right) +
\sum\limits_{i = 1}^M {KLD\left( {{f_i}\left( {{x_i};\eta _i^{(db)}} \right)||{g_i}\left( {{x_i};\eta _i^{(q)}} \right)} \right)}
\end{array}
\end{eqnarray}

\indent Now the second term of Eq.~\ref{eq22} corresponding to the closed KLD among the marginal functions should be calculated in order to obtain the overall closed form. According to \cite{Do}, KLD among two GG functions is calculated as in Eq.~\ref{eq23}, assuming $\mu  = 0$:

\begin{eqnarray}
\label{eq23}
\begin{array}{l}
KLD\left( {{f^{db}}\left( {{\alpha _{db}},{\beta _{db}}} \right)||{g^q}\left( {{\alpha _q},{\beta _q}} \right)} \right) = \\
 \log \left( {\frac{{{\beta _{db}}{\alpha _q}\Gamma \left( {{\raise0.7ex\hbox{$1$} \!\mathord{\left/
 {\vphantom {1 {{\beta _q}}}}\right.\kern-\nulldelimiterspace}
\!\lower0.7ex\hbox{${{\beta _q}}$}}} \right)}}{{{\beta _q}{\alpha _{db}}\Gamma \left( {{\raise0.7ex\hbox{$1$} \!\mathord{\left/
 {\vphantom {1 {{\beta _{db}}}}}\right.\kern-\nulldelimiterspace}
\!\lower0.7ex\hbox{${{\beta _{db}}}$}}} \right)}}} \right) 
+ {\left( {\frac{{{\alpha _{db}}}}{{{\alpha _q}}}} \right)^{{\beta _q}}}\frac{{\Gamma \left( {{\raise0.7ex\hbox{${\left( {{\beta _q} + 1} \right)}$} \!\mathord{\left/
 {\vphantom {{\left( {{\beta _q} + 1} \right)} {{\beta _{db}}}}}\right.\kern-\nulldelimiterspace}
\!\lower0.7ex\hbox{${{\beta _{db}}}$}}} \right)}}{{\Gamma \left( {{\raise0.7ex\hbox{$1$} \!\mathord{\left/
 {\vphantom {1 {{\beta _{db}}}}}\right.\kern-\nulldelimiterspace}
\!\lower0.7ex\hbox{${{\beta _{db}}}$}}} \right)}} - \frac{1}{{{\beta _{db}}}}
\end{array}
\end{eqnarray}

\indent Replacing Eq.~\ref{eq23} in Eq.~\ref{eq22}, the KLD closed form of GC-GG function is calculated as Eq.~\ref{eq24}:

\begin{eqnarray}
\label{eq24}
\begin{array}{l}
KLD\left( {f\left( {\overrightarrow x ;{\theta ^{db}}} \right)||g\left( {\overrightarrow x ;{\theta ^q}} \right)} \right) = \frac{1}{2}\left( {tr\left( {\Sigma _q^{ - 1}{\Sigma _{db}}} \right) + \log \frac{{\left| {{\Sigma _q}} \right|}}{{\left| {{\Sigma _{db}}} \right|}} - M} \right) + \\
{\rm{                                       }}\sum\limits_{i = 1}^M {\log \left( {\frac{{\beta _{db}^i\alpha _q^i\Gamma \left( {{\raise0.7ex\hbox{$1$} \!\mathord{\left/
 {\vphantom {1 {\beta _q^i}}}\right.\kern-\nulldelimiterspace}
\!\lower0.7ex\hbox{${\beta _q^i}$}}} \right)}}{{\beta _q^i\alpha _{db}^i\Gamma \left( {{\raise0.7ex\hbox{$1$} \!\mathord{\left/
 {\vphantom {1 {\beta _{db}^i}}}\right.\kern-\nulldelimiterspace}
\!\lower0.7ex\hbox{${\beta _{db}^i}$}}} \right)}}} \right) + {{\left( {\frac{{\alpha _{db}^i}}{{\alpha _q^i}}} \right)}^{\beta _q^i}}\frac{{\Gamma \left( {{\raise0.7ex\hbox{${\left( {\beta _q^i + 1} \right)}$} \!\mathord{\left/
 {\vphantom {{\left( {\beta _q^i + 1} \right)} {\beta _{db}^i}}}\right.\kern-\nulldelimiterspace}
\!\lower0.7ex\hbox{${\beta _{db}^i}$}}} \right)}}{{\Gamma \left( {{\raise0.7ex\hbox{$1$} \!\mathord{\left/
 {\vphantom {1 {\beta _{db}^i}}}\right.\kern-\nulldelimiterspace}
\!\lower0.7ex\hbox{${\beta _{db}^i}$}}} \right)}} - \frac{1}{{\beta _{db}^i}}} 
\end{array}
\end{eqnarray}

\indent Moreover, KLD between two TLS functions is calculated by Eq.~\ref{eq25}:

\begin{eqnarray}
\label{eq25}
\begin{array}{l}
KLD\left( {{f^{db}}\left( {{\mu _{db,}}{\alpha _{db}},{v_{db}}} \right)||{g^q}\left( {{\mu _{q,}}{\alpha _q},{v_q}} \right)} \right) = \\
\log \left( {\frac{{\frac{{\Gamma \left( {\frac{{{v_{db}} + 1}}{2}} \right)}}{{{\sigma _{db}}\sqrt {{v_{db}}\pi } \Gamma \left( {\frac{{{v_{db}}}}{2}} \right)}}}}{{\frac{{\Gamma \left( {\frac{{{v_q} + 1}}{2}} \right)}}{{{\sigma _q}\sqrt {{v_q}\pi } \Gamma \left( {\frac{{{v_q}}}{2}} \right)}}}}} \right) - \left( {\frac{{{v_{db}} + 1}}{2}} \right)\left( {\psi \left( {\frac{{{v_{db}} + 1}}{2}} \right) - \psi \left( {\frac{{{v_{db}}}}{2}} \right)} \right) + \\
 \left( {\frac{{{v_q} + 1}}{2}} \right)\left( {\psi \left( {\frac{{{v_q} + 1}}{2}} \right) - \psi \left( {\frac{{{v_q}}}{2}} \right)} \right)
\end{array}
\end{eqnarray}

\indent where $\psi \left( . \right)$ denotes the digamma function. Now by replacing Eq.~\ref{eq25} into Eq.~\ref{eq22}, the closed KLD for GC-TLS function is calculated by Eq.~\ref{eq26}:

\begin{eqnarray}
\label{eq26}
\begin{array}{l}
KLD\left( {f\left( {\overrightarrow x ;{\theta ^{db}}} \right)||g\left( {\overrightarrow x ;{\theta ^q}} \right)} \right) = \frac{1}{2}\left( {tr\left( {\Sigma _q^{ - 1}{\Sigma _{db}}} \right) + \log \frac{{\left| {{\Sigma _q}} \right|}}{{\left| {{\Sigma _{db}}} \right|}} - M} \right) + \\
{\rm{                                       }}\sum\limits_{i = 1}^M {\log \left( {\frac{{\frac{{\Gamma \left( {\frac{{v_{db}^i + 1}}{2}} \right)}}{{\sigma _{db}^i\sqrt {v_{db}^i\pi } \Gamma \left( {\frac{{v_{db}^i}}{2}} \right)}}}}{{\frac{{\Gamma \left( {\frac{{v_q^i + 1}}{2}} \right)}}{{\sigma _q^i\sqrt {v_q^i\pi } \Gamma \left( {\frac{{v_q^i}}{2}} \right)}}}}} \right) - \left( {\frac{{v_{db}^i + 1}}{2}} \right)\left( {\psi \left( {\frac{{v_{db}^i + 1}}{2}} \right) - \psi \left( {\frac{{v_{db}^i}}{2}} \right)} \right)} \\
 + \left( {\frac{{v_q^i + 1}}{2}} \right)\left( {\psi \left( {\frac{{v_q^i + 1}}{2}} \right) - \psi \left( {\frac{{v_q^i}}{2}} \right)} \right)
\end{array}
\end{eqnarray}

\section{Experiments and Results}
\indent In this section we first introduce the datasets and configurations used in the experiments and then compare our proposed method with the existing state-of-the-art methods. 

\subsection{Experiment Settings and The Evaluation Criterion}
\begin{figure*}[t!]
\begin{center}
\includegraphics[scale=.55]{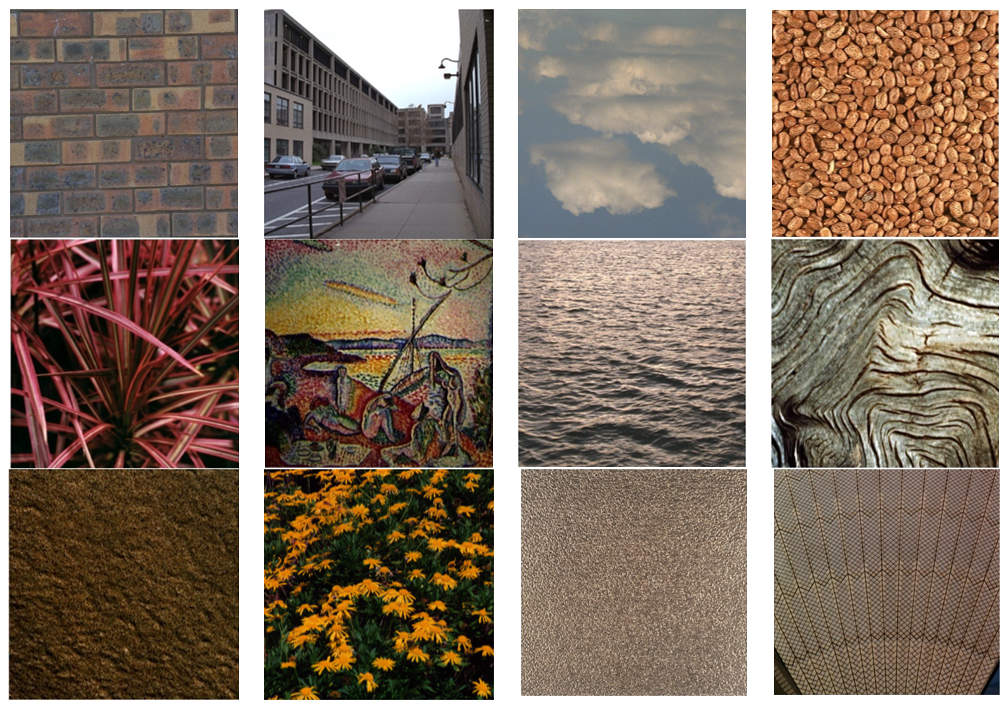}
\end{center}
\caption{\label{fig12} Samples of texture classes such as wall, building, cloud, food, wood, flower, painting, etc. in the VisTex dataset.}
\end{figure*}

\begin{figure*}[t!]
\begin{center}
\includegraphics[scale=.55]{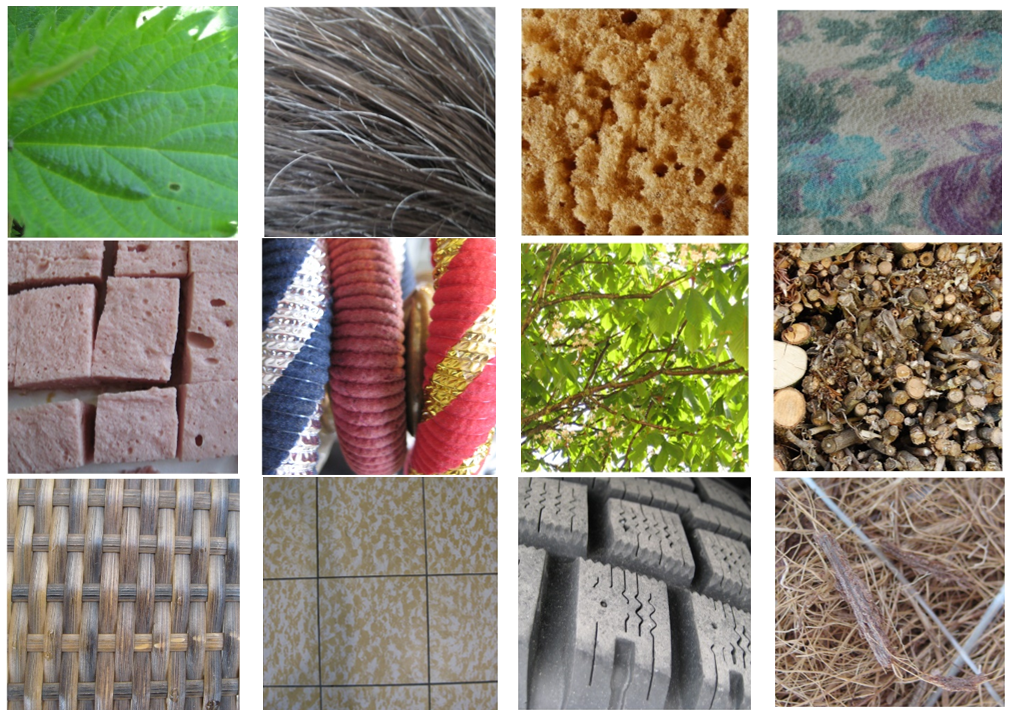}
\end{center}
\caption{\label{fig13} Samples of texture classes like leaves, hair, sponge, cloth, wood, tiles, tires, etc. in the STex dataset.}
\end{figure*}

\begin{figure*}[t!]
\begin{center}
\includegraphics[scale=.55]{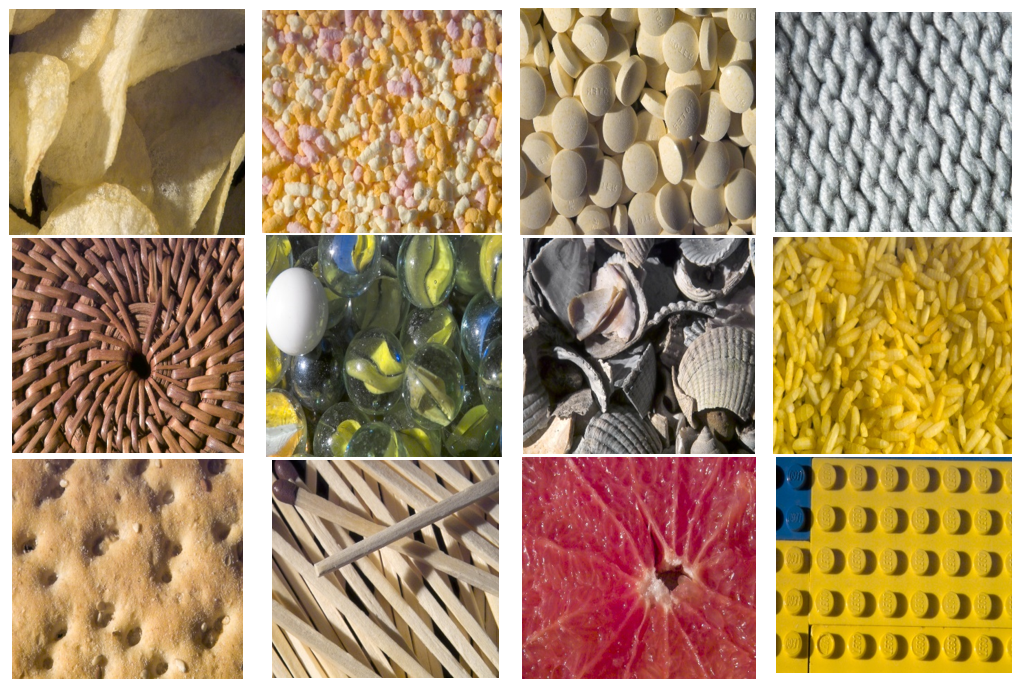}
\end{center}
\caption{\label{fig14} Samples of texture classes such as tablets, chips, rice, marbles, matches, biscuits, Legos, etc. in the ALOT dataset.}
\end{figure*}

\indent To perform the retrieval experiments of the proposed framework for color texture images, the four datasets VisTex(Full), VisTex(Small), ALOT, and STEX were used. The VisTex(Small) dataset contains 40 texture images chosen from the MIT Vision Texture Database, aimed at showing texture of real-world images. VisTex(Full) consists of 167 texture images from the same MIT database. These images were downloaded from MIT Vision Texture website \cite{Vistex}. Fig.~\ref{fig12} displays a few classes of the texture images in this dataset. There are 16 classes inside the VisTex(Small) dataset and 19 classes inside VisTex(Full). Another dataset used in the experiments, named STex (Salzburg Texture), belongs to University of Salzburg, Austria. This dataset consists of 476 separate texture images, including 32 different texture classes. Fig.~\ref{fig13} shows a few classes of texture images in STex dataset. The images of STex dataset were downloaded from University of Salzburg’s website \cite{STex}. The last dataset used in the experiments is ALOT, containing 250 different texture images taken under different illuminations. We have used images with $C1|1$ photo characteristic in our experiments. Fig.~\ref{fig14} presents a few classes of the texture images from ALOT dataset, obtained from \cite{ALOT}.

\indent The dimensions of images in these datasets are by default $512 \times 512$. Similar to many previous researches in the literature, we have divided the original images into 16 non-overlapping patches of dimensions $128 \times 128$. Therefore, the number of sub-images used in the evaluation are 640, 2672, 7616, and 4000 for the datasets VisTex(Small), VisTex(Full), STex, and ALOT, respectively.

\indent Average Retrieval Rate (ARR) was used to evaluate the proposed image retrieval framework. Since we have divided each image into 16 sub-images, the average percentage of the retrieved images belonging to the original image in the 16 retrieved sub-images is considered as ARR. Eq.~\ref{eq27} calculates the ARR of the sub-images: 

\begin{eqnarray}
\label{eq27}
ARR = \frac{1}{{{N_t}}}\sum\limits_{k = 1}^{{N_t}} {\frac{{{r_{{N_r}}}\left( {{I_k}} \right)}}{{{N_r}}}}
\end{eqnarray}

\indent where ${N_t}$ denotes the total number of sub-images in the dataset, ${I_k}$ represents the kth query sub-image, and ${r_{{N_r}}}\left( {{I_k}} \right)$ is the query function which represents the number of correctly retrieved sub-images corresponding to ${I_k}$  among the ${N_t}$ retrieved sub-images (i.e. sub-images that belong to a unified image along with the query image ${I_k}$).

\indent In these experiments we have used NSST with MaxFlat filters of size 8 in all scales in order to extract features. We have implemented the transform up to three decomposition scales, and have extracted 4, 8, and 16 directions in each of the scales, respectively.

\subsection{Performance Evaluation of The Proposed Retrieval Framework}
\begin{figure*}[t!]
\begin{center}
\includegraphics[scale=.55]{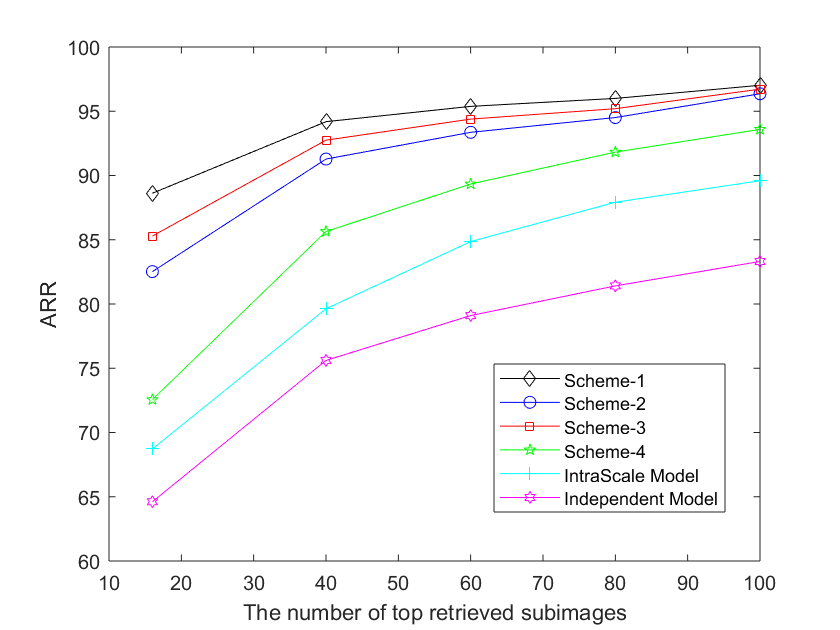}
\end{center}
\caption{\label{fig15} ARR values for five GC-GG models and one joint model assuming independence (GG) based on the changes in the number of retrieved sub-images.}
\end{figure*}

\begin{figure*}[t!]
\begin{center}
\includegraphics[scale=.85]{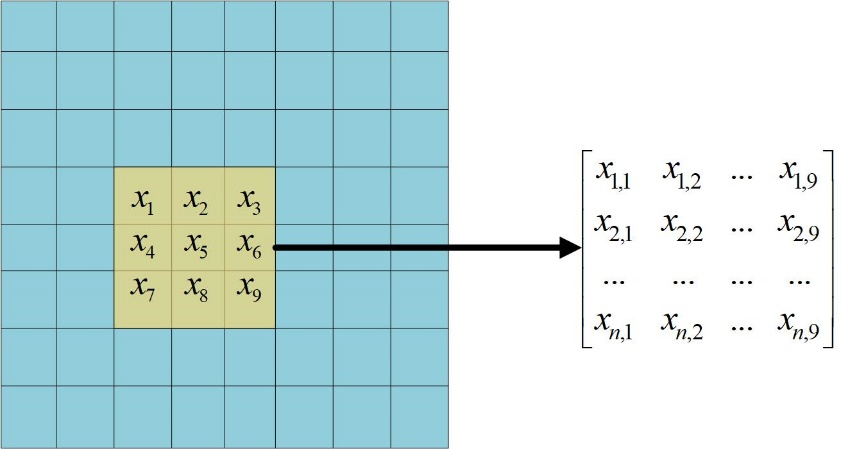}
\end{center}
\caption{\label{fig16} Creating the data matrix in the intra-scale model. The number of coefficients inside the sliding window is equal to the number of features in the matrix.}
\end{figure*}

\indent In order to show that the joint modeling of the coefficients using Copula function on the schematics in Fig.~\ref{fig8} in the texture image retrieval framework provides better ARR compared to the joint model with the assumption of independency, we have investigated six models including five types of joint Copula-based models and one joint model with the assumption of independency of NSST coefficients on the VisTex(Small) dataset. In Copula-based models, GC-GG is used for modeling. Fig.~\ref{fig15} shows the plot of ARR based on the changes in the number of retrieved images for the six aforementioned models. As can be seen from Fig.~\ref{fig15}, all joint models have a better ARR compared to the joint model with the assumption of independency. In general, it can be inferred that all of the proposed models have a robust performance to changes in the number of retrieved images. Moreover, schemes one, two and three have a better precision compared to scheme four (independence of color channels). We saw before that inter-channel dependencies of NSST coefficients, has higher intensity compared to other types of dependencies. \indent In this experiment, to create an inter-scale model, the reference coefficient and the coefficient corresponding to the neighbor with highest mutual information with the reference coefficient (the neighboring with a $3 \times 3$ window was assumed), were used. Fig.~\ref{fig16}, shows the data matrix provided by the $3 \times 3$ sliding window in the inter-scale model. As can be seen, the result is a 9-column matrix (9-tuple neighborhood).

After generating the data matrix, the neighbor with the highest MI value relative to the reference coefficient, is chosen as the final neighbor for modeling. ARR of the intra-scale model is better than the joint model assuming independence, but performs weaker than schemes one through four. Table~\ref{table1} presents the ARR values calculated for six different models on the VisTex(Small) dataset. The results in this table also confirm the results obtained from Fig.~\ref{fig15}. 

\begin{table}
\begin{center}
\caption{\label{table1} ARR values for five GC-GG models and one joint model assuming independence (GG) on the VisTex(Small) dataset.}
\begin{tabular}{c|c|c|c}
\hline \thead{Model} & \thead{Dependencies} & \thead{Independencies} & \thead{ARR(\%)}  \\
\hline
\hline
Scheme-1&All&---&97.02\\
\cline{1-4}
Scheme-2&Color, Direction&Scale&96.36\\
\cline{1-4}
Scheme-3&Color, Scale&Direction&96.72\\
\cline{1-4}
Scheme-4&Scale, Direction&Color&93.59\\
\cline{1-4}
Intra-Scale&Neighborhood&Scale, Direction, Color&89.59\\
\cline{1-4}
Independent&---&All&83.31\\
\hline \hline
\end{tabular}
\end{center}
\end{table}

\indent To compare the proposed method with the existing state-of-the-art models we have used the joint Copula-based model with the first scheme. Since the datasets used in the literature were similar to the introduced datasets, we extracted the ARR values directly from the articles available in the literature. Table~\ref{table2} provides the ARR values of the proposed model and the set of existing state-of-the-art methods for the STex, ALOT, VisTex(Full), and VisTex(Small) datasets. We have named some of the methods in the $X/Y$ format, where $X$ represents the name of the model along with the extracted feature and $Y$ is the name of similarity criterion used in the methods. For instance, our proposed method is named GC-GG NSST/JD or GC-TLS NSST/JD which means that NSST coefficients have been modeled with GC-GG or GC-TLS models, and were compared using the JD similarity criterion. Numbers in Table~\ref{table2} that are highlighted in bold, show the methods with the highest ARR values among the compared methods. As can be seen, our proposed GC-GG NSST/JD method obtains the highest ARR values in the three datasets ALOT, VisTex(Full) and VisTex(Small), and is among the three best methods for the dataset STex. 

\begin{table}
\begin{center}
\caption{\label{table2} ARR values of the proposed methods along with the existing state-of-the-art methods on the four STex, ALOT, VisTex(Full), and VisTex(Small) datasets.}
\begin{tabular}{c|c|c|c|c}
\hline \thead{Method} & \thead{STex} & \thead{ALOT} & \thead{VisTex(Full)} & \thead{VisTex(Small)}  \\
\hline
\hline
LBP Histogram/Chi-square\cite{Ojala}&54.89&39.57&50.20&85.84\\
\cline{1-5}
CLBP Histogram/Chi-square\cite{Guo}&58.40&49.60&56.90&89.40\\
\cline{1-5}
DWT Histogram/Chi-square\cite{Bai}&59.70&50.90&61.20&91.70\\
\cline{1-5}
CIF-LBP\cite{Liu}&40.95&70.69&-&96.09\\
\cline{1-5}
DWT-Gamma/KLD\cite{Choy}&52.90&40.7&-&90.43\\
\cline{1-5}
DTCWT Weibull/KLD\cite{Kwitt2008}&58.80&40.60&55.40&84.00\\
\cline{1-5}
Gaussian Copula-Weibull/ML\cite{Kwitt2011}&70.06&54.10&63.00&89.50\\
\cline{1-5}
Student’s t Copula-Gamma/ML\cite{Kwitt2011}&64.30&47.50&63.80&88.90\\
\cline{1-5}
Student’s t Copula-GG/ML\cite{Kwitt2011}&70.60&50.08&63.20&88.90\\
\cline{1-5}
MPE/Geodesic\cite{Verdoolaege}&71.30&49.30&69.30&91.20\\
\cline{1-5}
EMM/ML\cite{Vasconcelos}&73.70&53.00&67.70&88.90\\
\cline{1-5}
Gabor Wavelet-Copula/KLD\cite{Li2017}&76.40&60.80&66.10&92.40\\
\cline{1-5}
ODBTC\cite{Guo2014}&-&43.62&-&90.67\\
\cline{1-5}
EDBTC\cite{Guo20144}&-&-&-&90.09\\
\cline{1-5}
DDDBTC\cite{Guo2015}&44.79&48.64&-&92.65\\
\cline{1-5}
LECoP\cite{Verma}&74.15&-&-&92.99\\
\cline{1-5}
LED/RD\cite{Pham}&80.08&-&-&94.70\\
\cline{1-5}
OWT-MDCM\cite{Li2019}&72.28&48.13&-&91.08\\
\cline{1-5}
DTCWT-MDCM\cite{Li2019}&77.01&55.74&-&93.13\\
\cline{1-5}
Gabor-MDCM\cite{Li2019}&83.36&60.36&-&94.15\\
\cline{1-5}
Mwavelets-MDCM\cite{Li2019}&\textbf{85.46}&61.88&-&95.68\\
\cline{1-5}
GC-tLS NSST/JD&78.36&70.95&68.43&96.21\\
\cline{1-5}
GC-GG NSST/JD&80.81&\textbf{72.82}&\textbf{69.47}&\textbf{97.02}\\
\hline \hline
\end{tabular}
\end{center}
\end{table}

\indent The ARR value of the proposed method is compared with Convolutional Neural Network (CNN) based methods on three STex, ALOT, and VisTex(Small) datasets in Table~\ref{table3}. As can be seen from Table~\ref{table3}, our proposed statistical model has a better performance compared to other methods on the VisTex(Small) dataset that has a smaller number of samples. Moreover, our proposed method, along with the ResNet network, have the highest ARR values on ALOT dataset. In STex dataset, where the number of samples is higher compared to the other datasets, the proposed method has the highest ARR after the ResNet method. In general, the acceptable performance of the proposed method compared to the methods that extract features using CNN networks, is realized by looking at Table~\ref{table3}, because the CNN-based methods require more training and test time, as well as more hardware resources, compared to statistical methods. 

\begin{table}
\begin{center}
\caption{\label{table3} ARR values of the proposed methods and the existing convolutional neural network-based methods on STex, ALOT, and VisTex(Small) datasets.}
\begin{tabular}{c|c|c|c}
\hline \thead{Method} & \thead{STex} & \thead{ALOT} & \thead{VisTex(Small)}  \\
\hline
\hline
CNN-AlexNet\cite{Krizhevsky}&68.32&59.01&91.76\\
\cline{1-4}
CNN-VGG16\cite{Simonyan}&72.16&60.34&92.48\\
\cline{1-4}
CNN-VGG19\cite{Simonyan}&71.68&59.15&92.02\\
\cline{1-4}
GoogleNet\cite{Szegedy}&77.60&60.71&92.87\\
\cline{1-4}
ResNet101\cite{He}&91.18&75.60&95.91\\
\cline{1-4}
ResNet50\cite{He}&91.59&75.68&96.28\\
\cline{1-4}
GC-tLS NSST/JD&78.36&70.95&96.21\\
\cline{1-4}
GC-GG NSST/JD&80.81&72.82&97.02\\
\hline \hline
\end{tabular}
\end{center}
\end{table}

\indent Another important issue in examining the performance of retrieval frameworks, is the performance of retrieval time. Eq.~\ref{eq28} calculates the total retrieval time:

\begin{eqnarray}
\label{eq28}
{t_{Total}} = {t_{FE}} + {t_{SM}}
\end{eqnarray}

\indent where the total retrieval time ${t_{Total}}$ consists of feature extraction time ${t_{FE}}$ and similarity matching time ${t_{SM}}$. In content-based image retrieval systems, when the dataset is large, the similarity matching (SM) time gains more importance because feature extraction can be done in advance on the dataset. As mentioned earlier, using the closed form KLD similarity criterion increases the retrieval speed compared to the normal case of calculating KLD and ML-based methods. This is because we only need the model parameters in the closed form, while the traditional ML and KLD methods require all variables (coefficients). Therefore, we expect our proposed framework which uses the closed form of KLD for calculating JD, to have an acceptable similarity matching time. For example we have presented the retrieval times for our proposed GC-GG NSST/JD method on the VisTex(Small) dataset in Table~\ref{table4}, where all the implementations were performed using Matlab2018-b on a laptop with Corei7-8550 U 2.00GHz CPU and with 12GB of RAM.

\begin{table}
\begin{center}
\caption{\label{table4} Retrieval times of the proposed GC-GG NSST/JD frameworks on the VisTex(Small) dataset.}
\begin{tabular}{c|c|c|c|c|c|c|c}
\hline  
& \multicolumn{2}{c|}{FE Time(s)} & \multicolumn{2}{c|}{SM Time(s)} & 
\multicolumn{2}{c|}{Total Time(s)} & \\
\hline
\makecell{Scheme} & \thead{$t_{db}$} & \thead{$t_{image}$} & \thead{$t_{db}$} & \thead{$t_{image}$} & 
\thead{$t_{db}$} & \thead{$t_{image}$} & \thead{ARR(\%)} \\
\hline \hline
Scheme-1&109.73&0.1715&30.34&0.047&140.07&0.2185&97.02\\
\cline{1-8}
Scheme-2&96.18&0.1503&15.98&0.025&112.16&0.1753&96.24\\
\cline{1-8}
Scheme-3&98.25&0.1535&12.26&0.019&110.51&0.1725&96.72\\
\cline{1-8}
Scheme-4&100.47&0.1570&14.38&0.022&114.85&0.1790&93.59\\
\cline{1-8}
Intra-Scale&213.69&0.3339&8.28&0.012&221.97&0.3459&89.59\\
\cline{1-8}
Independent&96.12&0.1502&0.0022&0.000003&96.12&0.1502&83.31\\
\hline \hline
\end{tabular}
\end{center}
\end{table}

\indent Table~\ref{table4} covers total retrieval time, feature extraction (FE) time, and similarity matching (SM) time for different schemes of the GC-GG NSST/JD model for one image, ${t_{image}}$, and for all of the dataset, ${t_{db}}$. As expected, scheme one has longer retrieval time compared to schemes two to four. Furthermore, the reason for long feature extraction time in the intra-scale method, is the calculation of mutual information among the reference coefficient and its neighbors, because this must be done for all the sub-bands. Overall, the retrieval time of the scheme 1 of the proposed method on the VisTex(Small) dataset, is equal to 140 seconds, which displays the computational efficiency of the proposed method.

\section{Conclusion}
\indent We have presented a model based on Gaussian Copula in the shearlet domain in order to provide a color texture image retrieval framework. We introduced six different schemes for joint modeling of NSST sub-bands by defining four types of neighboring for the NSST sub-band coefficients. In the second step of retrieval, we used the JD (symmetric KLD) criterion to examine the similarities in our proposed framework, and derived its closed form for GC-GG, and GC-TLS models. Use of the closed form of KLD versus the conventional numerical computation of KLD, significantly increased the speed in the proposed retrieval framework. We investigated the performance of our proposed method against state-of-the-art methods, by using four well-known datasets in the field of color texture image retrieval. It was seen that the proposed method provided a better retrieval precision compared to the state-of-the-art statistical modeling methods and a precision equal to the best convolutional neural network-based methods. Moreover, we investigated retrieval speed of the proposed methods and found that using the closed form of KLD along with non-subsampled shearlet transform, creates a fast framework with an acceptable speed for color texture image retrieval.

\bibliographystyle{elsarticle-num}

\bibliography{mybibfile}

\let\mybibitem\bibitem

\end{document}